\documentclass[10pt,twocolumn,letterpaper]{article}

\usepackage[pagenumbers]{cvpr}

\usepackage{stfloats}

\usepackage{microtype}

\definecolor{cvprblue}{rgb}{0.21,0.49,0.74}
\usepackage[pagebackref,breaklinks,colorlinks,allcolors=cvprblue]{hyperref}

\title{Understanding and Repairing Cooperative Perception under Adverse Weather}

\author{
  Mengxu Ma, Linshan Yan, Kailong Zhang, Ruiyang Guo \\
  Dagang Shen, Yichen Fan, Hongping Gan \\[3pt]
  Northwestern Polytechnical University, Xi'an, China
}

\hypersetup{
  pdftitle={Understanding and Repairing Cooperative Perception under Adverse Weather},
  pdfauthor={Mengxu Ma, Linshan Yan, Kailong Zhang, Ruiyang Guo, Dagang Shen, Yichen Fan, Hongping Gan}
}


\renewcommand{\topfraction}{.94}

\renewcommand{\textfraction}{.06}
\renewcommand{\dbltopfraction}{.94}

\begin{document}
\maketitle
\raggedbottom
\begin{abstract}
A cooperative detector can miss a vehicle that it detected in the original scan, even though only the surrounding LiDAR returns have changed.
Such misses occur in all 16 V2X-R test scenes in both simulated fog and snow.
Based on these observations, we remove exterior points from CoInfra scans and hold the altered inputs fixed. Restoring V2X-ViT's pre-change features around all annotated objects just before fusion recovers 71/72 induced misses. Each of three equal-area exterior restorations recovers 0/72.
We trace these misses to changes in pre-fusion object-neighborhood representations, a mechanism we call Weather-Induced Context Interference.
Across V2X-R scenes combined, changes confined to vehicle boxes cause more misses than exterior changes alone in fog. In snow, exterior changes cause more misses.
Without pre-change features at inference, we examine the current candidates. Some missed targets retain an accurate box, while others have none that meets the overlap threshold. Adverse-Weather Cooperative Repair (AWCR) generates bounded alternatives to the original boxes. Original and corrected boxes compete for the same shortlist. AWCR jointly ranks the admitted boxes using current receiver and peer measurements.
Across three fits, AWCR raises V2X-ViT's car BEV AP@0.7 by 4.22 points on CoInfra adverse-weather development recordings and by 3.71 on a held-out heavy-snow recording. On natural adverse-weather inputs, reranking recovers overlooked accurate boxes, and local correction recovers some targets whose best original box narrowly misses the overlap threshold.
\end{abstract}

\section{Introduction}
\label{sec:intro}

\begin{figure}[t]
  \centering
  \includegraphics[width=\linewidth]{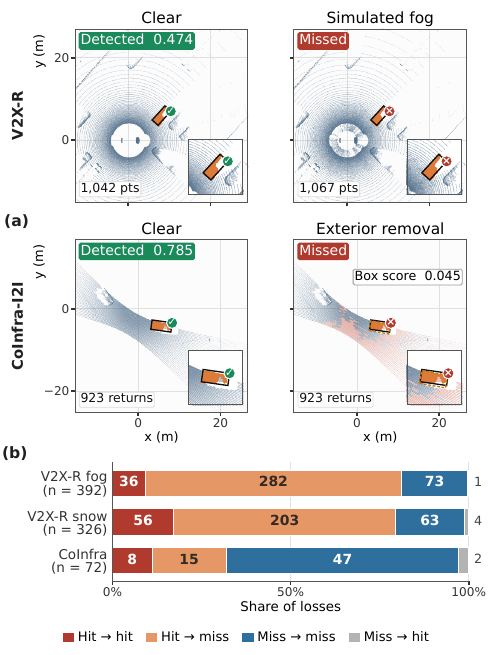}
  \caption{Cooperative misses. (a) Fog preserves all 1,042 original point locations in the BEV footprint and adds 25 points. Exterior removal preserves all 923 body returns and drops 9,879 others. The dashed box is the highest-scoring accurate candidate.
  (b) Targets detected cooperatively before the change and missed afterward are grouped by their receiver-only outcomes on the two inputs. The V2X-R bars show complete simulated weather at 72 timestamps. CoInfra covers 320 views at 80 timestamps. Table~\ref{tab:mechanism}(b) compares object-only and exterior-only weather changes on V2X-R.}
  \label{fig:problem_analysis}
\end{figure}

Cooperative perception combines LiDAR observations from different viewpoints to detect vehicles that one sensor alone can miss~\cite{xu2022v2xvit,hu2022where2comm}.
Weather changes returns from both vehicles and their surroundings. Figure~\ref{fig:problem_analysis}(a) shows a simulated fog miss with new points inside the vehicle footprint and a miss induced by removing only exterior returns from a real clear-weather scan.
When a vehicle's own returns stay fixed, how can changes in its surroundings make a cooperative detector miss it?

Prior work improves weather robustness through fusion and denoising~\cite{chae2024lidarradar,heinzler2020denoising,huang2025v2xr,yang2026hrcp}, tests cooperative detectors under corruption~\cite{du2025rcpbench,guo2024cootest}, and uses background context to guide communication~\cite{han2026colc,wu2026fadelead}.
We pair clear and simulated-weather V2X-R scans at the same instant. Replacing only exterior returns induces misses in all 16 test scenes in both fog and snow (Section~\ref{sec:exterior_intervention}).
Real recordings rarely provide such pairs. On copies of CoInfra scans, we remove exterior returns using directional loss patterns from adverse-weather training data.
This intervention induces 72 V2X-ViT misses. In 47 of them, the receiver alone misses the vehicle both before and after removal (Figure~\ref{fig:problem_analysis}(b)).

We keep the altered CoInfra inputs fixed and restore pre-change features just before fusion.
Restoring all annotated object neighborhoods recovers 71/72 induced misses. Each of three equal-area exterior restorations recovers 0/72 (Table~\ref{tab:mechanism}).
Neighborhood restoration recovers more misses than any exterior control in each V2X-R test scene and each of three V2X-Radar-C recordings under controlled removal. Most CoInfra misses persist even with the initial encodings of object-containing cells held fixed.
An exterior change can thus alter the representation around an object and turn a correct detection into a miss. We call this failure \emph{Weather-Induced Context Interference}.

The paired V2X-R scans also let us compare regions. Across scenes combined, changing only the returns inside vehicle boxes causes more misses in fog than changing only exterior returns. In snow, exterior changes cause more misses.

Without the pre-change features at inference, what can the detector recover from its current candidates?
Of the same 72 CoInfra misses, 31 retain an accurate candidate before final selection. The remaining 41 have no accurate original box, so recovery requires a change in geometry. Both candidate states also occur in unmodified natural adverse-weather scans.
These two candidate states motivate \emph{Adverse-Weather Cooperative Repair} (AWCR).
\emph{Local Correction} generates bounded alternatives to original boxes. \emph{Quality Selection} uses current receiver and peer measurements to rank originals and alternatives together. Reranking can recover an accurate original discarded by thresholding or suppression. For some natural misses, a locally corrected box reaches the required overlap when no original box does.

Across three fits, AWCR raises V2X-ViT car BEV AP@0.7 by 4.22 points on CoInfra adverse-weather development inputs~\cite{ning2025coinfra} and by 3.71 on a held-out heavy-snow recording. On a rainy DAIR/V2X-Seq recording~\cite{yu2022dairv2x,yu2023v2xseq}, F-Cooper gains 5.00 vehicle AP@0.7 points but trails the receiver alone. On the full synchronous V2X-Radar-C validation set, HEAL with AWCR reaches 68.22 vehicle BEV AP@0.7, exceeding the best LiDAR-only result in the official benchmark (67.57)~\cite{yang2025v2xradar}.

In a separate V2X-R experiment, we use weather-specific simulator point labels to train a \emph{Point Classifier} for each weather. The known test weather selects the classifier that filters peer returns (Figure~\ref{fig:framework}(c)). Under this protocol, vehicle BEV AP@0.7 reaches 68.34 for fog-trained AttFuse in fog and 55.49 for CoAlign in snow (Table~\ref{tab:supp_filtering_public}).

We make three contributions.
\begin{itemize}
  \item With vehicle returns held fixed, we trace misses caused by exterior changes to the pre-fusion features around the target.
  \item We introduce AWCR, which keeps original boxes eligible alongside bounded corrections and uses current receiver and peer returns to rank both sets before suppression.
  \item With AWCR, V2X-ViT gains 3.71 car AP70 points on CoInfra's held-out heavy-snow recording and HEAL reaches 68.22 vehicle AP70 on V2X-Radar-C. We track recoveries by target and repair step (Section~\ref{sec:component_roles}).
\end{itemize}

\section{Related Work}
\label{sec:related_work}

\subsection{Weather Robustness and Context}
V2X-R uses radar features to guide LiDAR denoising in adverse weather~\cite{huang2025v2xr}, while V2X-DGW studies generalization across weather conditions~\cite{li2025v2xdgw}.
Under bandwidth constraints, CoLC samples both foreground and background points and completes the cloud at the receiver~\cite{han2026colc}. FadeLead learns to encode background context in the foreground features it transmits~\cite{wu2026fadelead}.

\subsection{Diagnosing Cooperative Detection Failures}
CooTest identifies targets that a receiver detects alone but misses or localizes incorrectly under cooperation~\cite{guo2024cootest}. RCP-Bench varies which agents experience camera corruptions to measure the impact of corrupted collaborators on detection~\cite{du2025rcpbench}.
We compare cooperative detections of the same target before and after changing only the surrounding LiDAR returns. This includes detections that originally depended on cooperation.
We test the role of object-neighborhood features by comparing which misses are recovered when pre-change features are restored there or in equal-area exterior regions immediately before fusion.

\subsection{Candidate Refinement and Selection}
PV-RCNN refines proposals using keypoint features~\cite{shi2020pvrcnn}. From Voxel to Point combines point, BEV, and box-corner features to estimate the localization quality of refined boxes~\cite{li2021voxelpoint}.
For boxes received from peers, mmCooper predicts bounded quality scores and positional offsets from fused features. It discards low-quality boxes, corrects the remaining boxes, and merges them with the receiver's detections~\cite{liu2025mmcooper}.

AWCR operates on a fixed cooperative detector's own candidates before final selection. Bounded adjustments to box parameters and scores produce alternatives alongside the originals.
To rank both sets before suppression, AWCR estimates localization quality from box attributes and local point statistics measured separately in the receiver and peer scans. This leaves an accurate original box eligible even if its correction is poorly supported.

\section{Weather-Induced Context Interference}
\label{sec:mechanism}

\begin{figure*}[t]
  \centering
  \includegraphics[width=\textwidth]{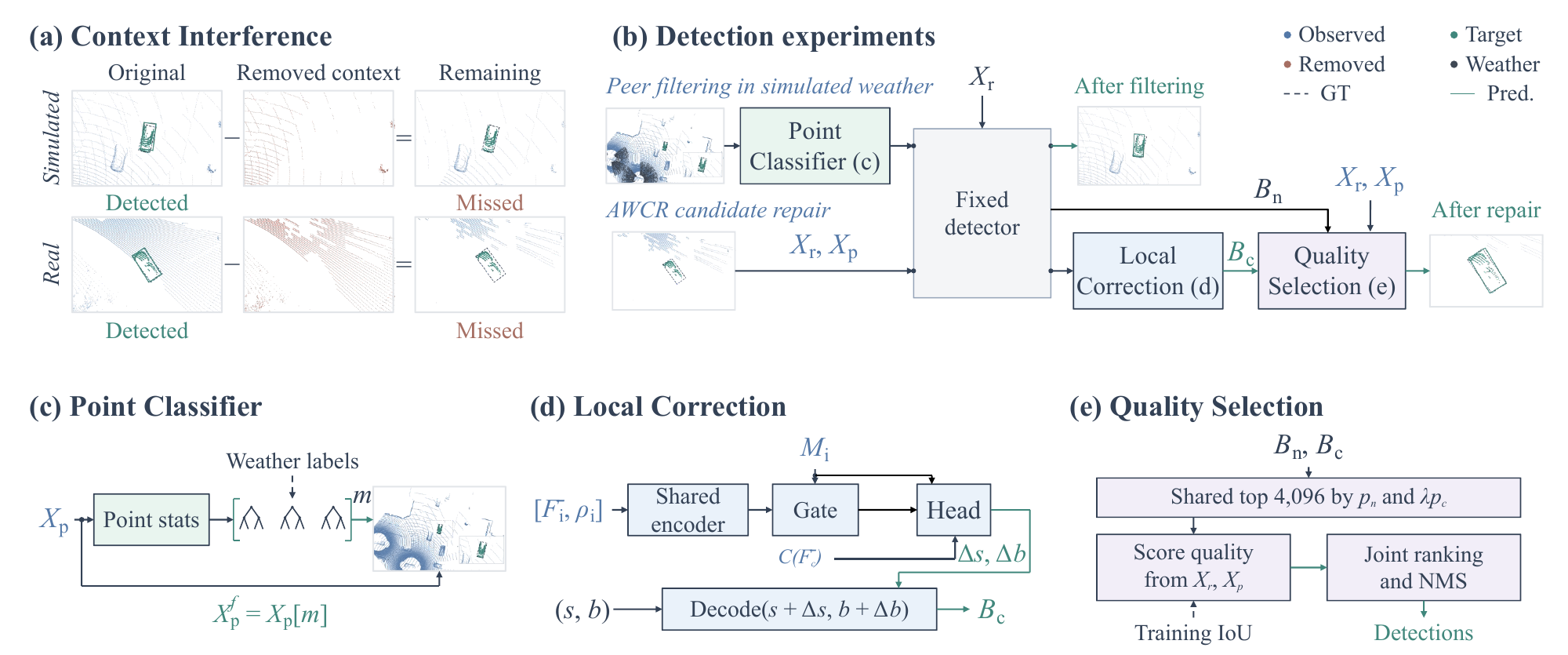}
  \caption{Exterior interference, AWCR candidate repair, and a separate point-filtering experiment.
  (a) Exterior removal under fixed angular masks causes misses while preserving 381 simulated and 316 real target returns. The real row starts from a clear input.
  (b) The upper row shows peer point filtering in simulated fog. The lower row applies AWCR candidate repair to the changed real input from (a).
  (c) Point Classifier uses generated-point identities as training labels.
  (d) Local Correction generates bounded alternatives.
  (e) Both branches share 4,096 slots. Quality Selection scores admitted boxes for joint ranking and class-wise NMS (Section~\ref{sec:quality_selection}).
  $X_r$ and $X_p$ denote receiver and peer point clouds, and $B_n$ and $B_c$ denote original and corrected candidates.
  GT denotes ground truth. Dashed arrows indicate training supervision.}
  \label{fig:framework}
\end{figure*}

\subsection{Changing Object and Exterior Returns}
\label{sec:exterior_intervention}
To separate weather changes on vehicles from those in their surroundings, we pair each clear V2X-R scan with simulated fog and snow at the same instant. Across all 16 test scenes and 3,166 times (3,133 with a peer), we replace the returns either inside annotated vehicle boxes or outside them. Exterior replacement preserves the coordinates, intensities, and retained points inside the boxes (Section~\ref{supp:weather_counterfactual}).

On CoInfra, we isolate the exterior effect by removing points outside annotated objects in all 320 development views at 80 clear and adverse-weather times (Figure~\ref{fig:framework}(a)). Each source's fixed angular mask covers directions with persistently fewer returns in adverse-weather training observations than in clear ones. Random removal deletes the same number of exterior points per source. Detector weights stay fixed throughout.

We track each target at one time in one receiving view as an observation.
A detection counts as correct when its score exceeds 0.27 and its rotated ground-plane overlap reaches 0.7 under class-aware matching.
A cooperative benefit is a correctly detected target that is missed by the receiver alone under the same rule.

\begin{table}[t]
  \centering
  \caption{Regional changes to LiDAR returns and pre-fusion feature restoration.
  (a) Lost/gained correct observations under directional and equal-count random removal. Random ranges use three fixed seeds for CoInfra and one for each public recording. Public~1--3 use F-Cooper (Section~\ref{supp:public_mechanism}).
  (b) Misses follow changes to either object or exterior returns (exterior only on CoInfra). Recoveries count exterior-induced misses found by restoring all object neighborhoods or each of three equal-area exterior controls (range). New losses count previously stable detections lost after object restoration.}
  \label{tab:mechanism}
  \begin{minipage}[t]{\columnwidth}
    \centering
    \small
    (a) Exterior removal\par\smallskip
    \setlength{\tabcolsep}{3pt}
    \begin{tabular}{@{}lrrrrr@{}}
      \toprule
      & Original & \multicolumn{2}{c}{Directional} & \multicolumn{2}{c}{Random} \\
      Setting / fusion & correct & Lost & Gained & Lost & Gained \\
      \midrule
      CoInfra / V2X-ViT & 326 & 72 & 18 & 10--18 & 10--19 \\
      \midrule
      Public 1 / F-Cooper & 746 & 23 & 28 & 36 & 67 \\
      Public 2 / F-Cooper & 1,067 & 252 & 69 & 40 & 50 \\
      Public 3 / F-Cooper & 180 & 12 & 18 & 7 & 5 \\
      \bottomrule
    \end{tabular}
  \end{minipage}\par\medskip
  \begin{minipage}[t]{\columnwidth}
    \centering
    \small
    (b) Regional changes and restoration (V2X-ViT)\par\smallskip
    \setlength{\tabcolsep}{2.5pt}
    \begin{tabular}{@{}lrrrrr@{}}
      \toprule
      & \multicolumn{2}{c}{Induced misses} & \multicolumn{2}{c}{Recovered} & New \\
      Setting & Object & Exterior & Object & Exterior & losses \\
      \midrule
      CoInfra removal & --- & 72 & 71 & 0 & 0 \\
      V2X-R fog & 9,703 & 5,774 & 5,582 & 6--24 & 102 \\
      V2X-R snow & 4,739 & 10,092 & 9,950 & 8--14 & 37 \\
      \bottomrule
    \end{tabular}
  \end{minipage}
\end{table}

Among 22,642 V2X-R target observations correctly detected in clear weather, fog causes more misses when only object returns change (9,703) than when only exterior returns change (5,774). In snow, exterior changes cause more misses (10,092 versus 4,739). Exterior changes induce misses in every scene under both conditions (Tables~\ref{tab:mechanism}(b) and~\ref{tab:supp_weather_counterfactual}).

On CoInfra, directional removal turns 72 of V2X-ViT's 326 originally correct observations into misses. Equal-count random removal loses 10--18 across three seeds (Table~\ref{tab:mechanism}(a)).
These 72 misses span 39 times and both roadside pairs. In 47 cases the receiver alone misses the target both before and after removal, so cooperation had supplied the lost detection (Figure~\ref{fig:problem_analysis}(b)).
Among the 90 adverse-weather observations initially correct in both modes, each mode loses 14 (Section~\ref{supp:receiver_cooperation}).

AttFuse in simulation and F-Cooper~\cite{chen2019fcooper} on three public V2X-Radar-C recordings~\cite{yang2025v2xradar} also lose detections under angular-mask removal. On the public recordings, directional removal does not consistently cause more losses than random removal (Section~\ref{supp:interventions}).

\subsection{Restoring Features around Objects}
\label{sec:representation_intervention}

\begin{figure*}[t]
\centering
\includegraphics[width=.76\textwidth,trim=0bp 28bp 0bp 30bp,clip]{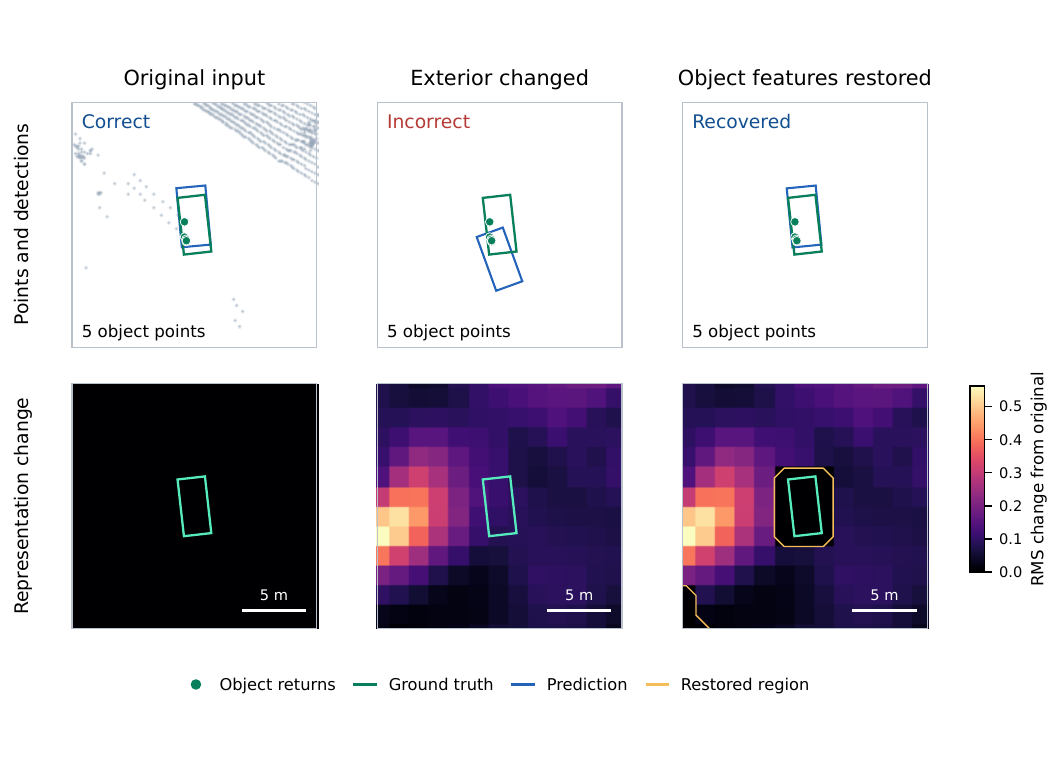}
\caption{Restoring object-neighborhood features after exterior removal. The top row shows predictions and retained points, including five unchanged target returns. The bottom row shows RMS change from original fusion-input features over both sources and 256 content channels on one color scale. Orange marks restored regions in the $20\times20$\,m crop. In this diagnostic, the middle and right columns use the same changed input points. The right column restores the original pre-fusion features in all annotated object neighborhoods.}
\label{fig:supp_representation_restoration}
\end{figure*}

Which pre-fusion features account for the 72 CoInfra misses? We keep the changed point clouds and weights fixed and restore selected features from the same scan before deletion (Figure~\ref{fig:supp_representation_restoration}).
Restoring the pre-change features in all annotated object neighborhoods together recovers 71/72 misses and preserves all 254 detections that survived removal. Restoring each of three predeclared equal-area exterior regions recovers 0/72 (Table~\ref{tab:mechanism}(b)).
The controls have similar mean range but different occupancy. The restoration contrast holds for both roadside pairs and separately on clear and adverse-weather inputs (Section~\ref{supp:restoration}).

An object return can share a pillar with a removed exterior point. In a stricter test, we protect every pillar containing an object return, preserving both its input points and its initial encoding.
Of the original 72 misses, 68 persist. Three other originally correct observations become misses, giving 71 induced misses under this condition.
Restoring pre-fusion features one target neighborhood at a time recovers 70/71 misses in this stricter test. Equal-area exterior restoration recovers none (Section~\ref{supp:encoding_control}).
Thus, exterior changes can affect detection even downstream of intact target-cell encodings.

On V2X-Radar-C (84 times, 168 views), F-Cooper's object-neighborhood restoration recovers 214/287 induced misses versus 11, 2, and 15 for the exterior controls. This contrast holds in all three recordings, although the second contributes 252/287 misses. Their weather is unverified (Section~\ref{supp:public_mechanism}).
On the full V2X-R test split, object-neighborhood restoration recovers 96.7\% of exterior-induced misses in fog and 98.6\% in snow. Each exterior control recovers fewer than 0.5\% in either weather (Table~\ref{tab:mechanism}(b)). Further controls cover density-robust training and DAIR/V2X-Seq (Sections~\ref{supp:robust_mechanism} and~\ref{supp:fcooper_restoration}).
Restoration locates the effect of the intervention. At inference, only the current observations and their candidates are available.

\subsection{Candidate Analysis for Repair}
\label{sec:candidate_diagnosis}

For the same 72 CoInfra misses, we rerun post-processing after restoring pre-removal values across the output grid. Box coordinates alone suffice in 37 cases, scores alone in 16, and 19 require both (Table~\ref{tab:supp_decision_decomposition}).

Which targets still have an accurate box on the changed input? We inspect all valid original car candidates before applying the repair budget of 4,096. For 31 misses, at least one box reaches 0.7 BEV IoU with the target. For the other 41, no box does, so reranking alone cannot recover them (Section~\ref{supp:full_candidate_evidence}).

On unmodified natural adverse-weather inputs, 117 of V2X-ViT's 432 misses retain an accurate cooperative candidate. In 108 of those 117 cases, every accurate box falls below the score threshold. In the remaining 9, non-maximum suppression removes the accurate boxes that pass the threshold (Section~\ref{supp:natural_evidence}).
Accurate candidates also remain in 42 of 168 clear-weather misses. On natural inputs, these counts identify opportunities to improve selection.

These observations motivate two tests. Can current measurements help select an accurate box that was discarded? When every original box falls short, how often can a bounded correction supply an accurate one? Section~\ref{sec:component_roles} tests both on controlled and natural inputs.

\section{Adverse-Weather Cooperative Repair}
\label{sec:awcr}

AWCR operates on the candidates identified in Section~\ref{sec:candidate_diagnosis}. Local Correction proposes bounded changes to box coordinates and scores from each anchor. Quality Selection ranks the original and corrected boxes together using current receiver and peer measurements, keeping accurate originals eligible for selection (Figure~\ref{fig:framework}(b,d,e)).

\subsection{Candidate Repair from Current Observations}
Given the receiver and peer clouds $X_r,X_p$, the fixed detector produces anchor logits $s$ and encoded box regressions $b$ for the repaired class. Decoding these outputs gives the original boxes $B_n=\operatorname{Decode}(s,b)$ with confidence $p_n=\sigma(s)$, where $\sigma$ is the sigmoid.

\subsection{Local Correction}
\label{sec:candidate_repair}
\label{sec:local_correction}
For each anchor, Local Correction uses local evidence from both sources and fused detector features to propose an alternative box and its initial confidence.
We align the points and shallow features from source $i\in\{r,p\}$ to the receiver frame. Within a $5\times5$ neighborhood, we pool aligned features over observed cells into $F_i$, then normalize by training-set root-mean-square scales to obtain $\bar F_i$. The same neighborhood provides normalized point density $\rho_i$ and a support mask $M_i$ (Section~\ref{supp:local_training}). At each supported cell, we concatenate four copies of the 64-channel $\bar F_i$ and one density channel to form the 257-channel input to the shared encoder $E$. A learned gate weights each cell's encoding.
\begin{equation}
  \begin{aligned}
    z_i &= M_i E([\bar F_i,\bar F_i,\bar F_i,\bar F_i,\rho_i]),\\
    h_i &= M_i\sigma(G(z_i))z_i.
  \end{aligned}
  \label{eq:supported_encoding}
\end{equation}

Both encodings have 64 channels. We likewise normalize the detector's fused context $F_c$ to $\bar F_c$. The head $H$ combines its 64-channel projection $C(\bar F_c)$ with the gated encodings and support masks (Figure~\ref{fig:framework}(d)). Applying $\tanh$ bounds the predicted residuals.
\begin{equation}
  \begin{aligned}
    (u_s,u_b)&=H([h_r,h_p,C(\bar F_c),M_r,M_p]),\\
    \Delta s&=3\tanh u_s,\qquad
    \Delta b=\beta\odot\tanh u_b.
  \end{aligned}
  \label{eq:bounded_residual}
\end{equation}
The elementwise product with $\beta$ limits center shifts to 0.5\,m per horizontal axis and 0.25\,m vertically, log-size changes to 0.12, and yaw changes to 0.2 radians (Eq.~\eqref{eq:reg_bounds}).
Adding the residuals to the same anchor outputs gives the corrected boxes
\begin{equation}
  B_c=\operatorname{Decode}(s+\Delta s,\ b+\Delta b).
  \label{eq:corrected_candidates}
\end{equation}
Their initial confidence is $p_c=\sigma(s+\Delta s)$. Both $B_n$ and $B_c$ remain available for selection.

\subsection{Quality Selection}
\label{sec:quality_selection}
For each input, original and corrected boxes of the repaired class share a single budget of 4,096 candidates (Figure~\ref{fig:framework}(e)). We pool both branches and admit the boxes with the highest priorities, using $p_n$ for original boxes and $\lambda p_c$ for corrected boxes, with $\lambda\in[0,1]$. Only admitted boxes proceed to quality scoring, with coordinates fixed.

For each admitted box $B$, we form a descriptor $v_B\in\mathbb{R}^{67}$ from seven confidence, geometry, and branch attributes and 30 local point statistics from each source. The confidence feature uses the logit of the branch's unweighted confidence $p$, either $p_n$ or $p_c$ (Section~\ref{supp:quality_training}).
We estimate localization quality from this descriptor with two gradient-boosting regressors. We fit $R_0$ on original boxes and $R_1$ on both branches. Each predicts the maximum BEV IoU with a visible same-class annotation. We use predicted overlap to adjust confidence~\cite{li2021voxelpoint}, clipping the estimate to $[0,1]$ before adding a bounded offset to the incoming logit.
\begin{equation}
  \begin{aligned}
    q_j(B)&=\operatorname{clip}(R_j(v_B),0,1),\\
    Q_j(B)&=\sigma\!\left(\operatorname{logit}(p)+2q_j(B)-1\right).
  \end{aligned}
  \label{eq:quality_score}
\end{equation}
The same $\lambda$ controls final scoring by blending the two quality-adjusted scores for each original box and scaling the joint regressor's score for each corrected box.
\begin{equation}
 p'_B=\begin{cases}
 (1-\lambda)Q_0(B)+\lambda Q_1(B),& B\in B_n,\\
 \lambda Q_1(B),& B\in B_c.
 \end{cases}
 \label{eq:correction_trust}
\end{equation}

The detector ranks original and corrected boxes together by $p'_B$ and applies class-wise non-maximum suppression (NMS) to the combined set. Overlapping boxes from either branch can suppress one another. In the natural CoInfra comparisons, the shared foreground-support filter is applied after scoring and before NMS (Section~\ref{supp:evaluation}).

At $\lambda=0$, only original boxes enter and receive $Q_0$. At $\lambda=1$, admission uses unweighted confidence and both branches use $Q_1$. Validation within CoInfra repair training selects $\lambda=0.7$ for the main comparisons, including the public-detector tests (Section~\ref{supp:correction_trust_selection}).

\begin{figure*}[!t]
\centering
\begin{minipage}{\textwidth}
  \centering
  {\makeatletter
  \let\Gread@transgrouptrue\Gread@transgroupfalse
  \makeatother
  \includegraphics[width=.94\textwidth,trim=0bp 4bp 0bp 14bp,clip]{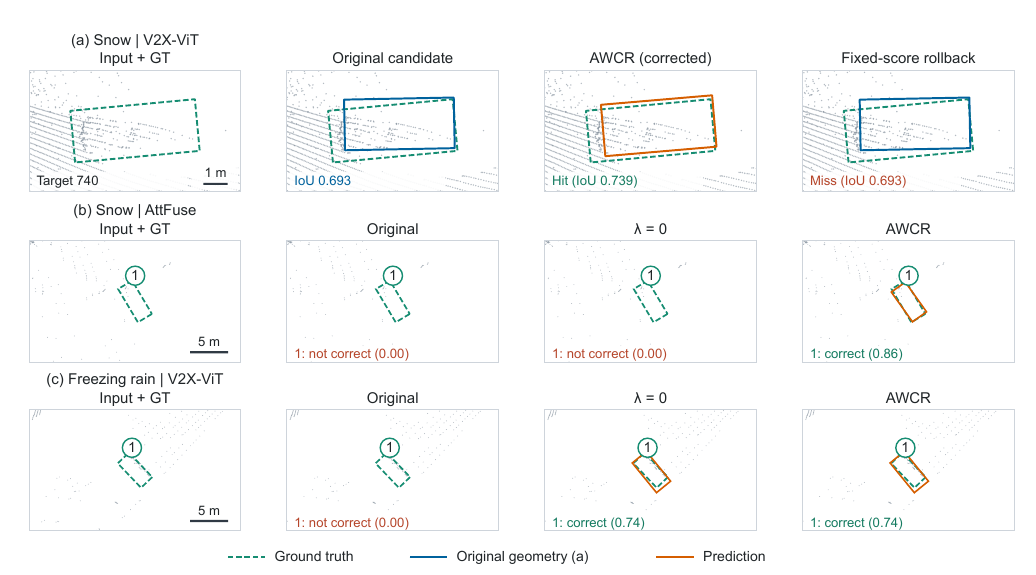}}
  \captionof{figure}{Natural adverse-weather CoInfra examples (seed 20261019). Dashed green boxes mark annotations.
  (a) Snow target 740 ($8.75\times5$\,m crop). Its best original box (blue, BEV IoU 0.693) and the corrected box (orange, 0.739) share an anchor. Reverting the coordinates at fixed score restores the miss.
  (b,c) Each $28\times16$\,m crop compares the detector, selection of original boxes ($\lambda=0$), and AWCR ($\lambda=0.7$). Orange shows retained car predictions with score $>0.27$ intersecting the crop. Parentheses give the marked target's best retained IoU (0 for no overlap). In (c), reranking alone recovers the target.}
  \label{fig:qualitative}
\end{minipage}
\end{figure*}

\subsection{Candidate Repair Training and Inference}
\label{sec:repair_supervision}
Visible annotations define the initial positive anchors. For CoInfra training variants with only exterior points removed, matched native detections on the clear counterpart add positive anchors with annotation-derived box targets.
For eligible boxes already accurate on the current input, the native regression serves as the target.
We mask classification loss for ambiguous negatives and positives that fail the bounded-update check (Section~\ref{supp:local_training}).
The correction network minimizes
\begin{equation}
  \mathcal L_{\mathrm{local}}=
  \mathcal L_{\mathrm{focal}}+2\mathcal L_{\mathrm{box}}
  +0.1\mathcal L_{\mathrm{keep}}.
  \label{eq:local_loss}
\end{equation}
$\mathcal L_{\mathrm{focal}}$ is focal classification loss~\cite{lin2017focal}, and $\mathcal L_{\mathrm{box}}$ is positive-anchor Smooth L1 loss with a sine heading error.
At originally correct training anchors, $\mathcal L_{\mathrm{keep}}$ penalizes probability drops beyond 0.05 and regression drift normalized by $\beta$.
We also zero-initialize the final layer of $H$ so the corrected candidates match the original predictions at the start of training.

We fit both quality regressors by squared error to $q_B^\star=\max_{g\in\mathcal G_{\mathrm{train}}}\operatorname{IoU}_{\mathrm{BEV}}(B,g)$ over visible annotations of the repaired class in the current input, using zero for an empty set. At inference, candidate repair uses current source features, detector outputs, and point statistics computed centrally from the retained points.

\subsection{Independent Point-Filtering Experiment}
On V2X-R, we separately test peer filtering in simulated weather (Figure~\ref{fig:framework}(c)). We train fog and snow point classifiers with weather-specific simulator labels for original and weather-added returns. At test time, the known weather selects the Point Classifier, which filters peer returns after native preprocessing and before feature encoding. We retain the full preprocessed receiver cloud to isolate the effect on detection (Section~\ref{supp:point_filtering}).

\section{Experiments}
\label{sec:experiments}

\begin{table*}[!t]
\centering
\caption{BEV AP70 ($\times100$) before and after (a) AWCR candidate repair or (b) peer point filtering. CoInfra uses the same two-source support filter for all Car predictions, including boxes predicted by the receiver alone. Other datasets evaluate Vehicle. In (a), V2X-DGW reports released frame order / global confidence ranking on identical predictions. Other rows use global ranking. HEAL uses $\lambda=0.7$ from CoInfra. Paired gains precede rounding. Bold marks the higher score under each evaluator. Scores with AWCR or filtering are three-fit means$\pm$sample SD ($\dagger$ one fit). Dashes indicate untested combinations.}
\label{tab:detection}\label{tab:public_detectors}\label{tab:detection_pairs}\label{tab:detection_references}
\small
\setlength{\tabcolsep}{4pt}
\begin{tabular}{@{}lllrrr@{}}
\toprule
\multicolumn{6}{@{}l}{\textit{(a) AWCR candidate repair}} \\
Dataset & Condition & Detector / configuration & Baseline & With AWCR & $\Delta$AP70 \\
\midrule
CoInfra & Adverse dev. & Receiver-only & 8.10 & $-$ & $-$ \\
& Adverse dev. & Late fusion & 14.73 & $-$ & $-$ \\
& Adverse dev. & V2X-ViT & 25.70 & \textbf{29.92}$\pm$0.34 & $+4.22$ \\
& Adverse dev. & V2X-ViT + DROR~\cite{charron2018dror} & 22.09 & $-$ & $-$ \\
& Adverse dev. & V2X-ViT + DSOR~\cite{kurup2021dsor} & 9.76 & $-$ & $-$ \\
& Adverse dev. & V2X-ViT (ordinary fine-tuning) & 25.16 & $-$ & $-$ \\
& Adverse dev. & V2X-ViT (RCP-Mix, LiDAR)~\cite{du2025rcpbench} & 27.55 & $-$ & $-$ \\
& Adverse dev. & V2X-ViT (Robo3D-DIT adapt.)~\cite{kong2023robo3d} & 29.11 & \textbf{31.47}$\pm$0.14 & $+2.36$ \\
& Adverse dev. & AttFuse & 10.55 & \textbf{15.84}$^\dagger$ & $+5.28$ \\
& Nov.\ 27 snow & V2X-ViT & 23.42 & \textbf{27.13}$\pm$0.17 & $+3.71$ \\
\midrule
V2X-Radar-C & Sync. val. & HEAL & 67.53 & \textbf{68.22}$\pm$0.04 & $+0.69$ \\
DAIR-V2X-C & Validation & FadeLead hybrid & 66.53 & \textbf{67.24}$\pm$0.04 & $+0.71$ \\
OPV2V-W & Fog & V2X-DGW & \textbf{65.58} / 70.06 & 62.78$\pm$0.21 / \textbf{70.41}$\pm$0.05 & $-2.79$ / $+0.35$ \\
& Rain & V2X-DGW & \textbf{69.33} / 75.44 & 65.36$\pm$0.15 / \textbf{75.63}$\pm$0.04 & $-3.98$ / $+0.18$ \\
& Snow & V2X-DGW & \textbf{59.56} / 63.48 & 55.91$\pm$0.30 / \textbf{63.79}$\pm$0.12 & $-3.65$ / $+0.31$ \\
\midrule
DAIR/V2X-Seq & Rain recording & F-Cooper & 32.86 & \textbf{37.86}$\pm$0.97 & $+5.00$ \\
& Dry recording & F-Cooper & 23.45 & \textbf{28.40}$\pm$0.65 & $+4.95$ \\
\midrule
\multicolumn{6}{@{}l}{\textit{(b) Peer point filtering in simulated weather (separate experiment, V2X-R frame order)}} \\
Dataset & Condition & Detector / configuration & Baseline & With point filtering & $\Delta$AP70 \\
\midrule
V2X-R & Fog & CoAlign & 44.36 & \textbf{50.64}$\pm$0.07 & $+6.27$ \\
& Snow & CoAlign & 38.93 & \textbf{55.49}$\pm$0.05 & $+16.55$ \\
& Fog & AttFuse (clean-trained) & 38.34 & \textbf{51.52}$\pm$0.17 & $+13.18$ \\
& Snow & AttFuse (clean-trained) & 30.51 & \textbf{50.70}$\pm$0.04 & $+20.18$ \\
& Fog & AttFuse (fog-trained) & 61.49 & \textbf{68.34}$\pm$0.04 & $+6.86$ \\
& Snow & AttFuse (fog-trained) & 45.74 & \textbf{54.17}$\pm$0.06 & $+8.43$ \\
\bottomrule
\end{tabular}
\end{table*}

\begin{table}[!t]
\centering
\caption{Quality Selection with fixed V2X-ViT candidates on natural adverse-weather CoInfra inputs. AP70 is multiplied by 100. Kept follows the same 230 originally correct observations, and False counts unmatched predictions above score 0.27. Learned rows use $\lambda=0$ and report three-fit mean AP with sample SD and mean counts. Learned variants share training samples, and all rows share candidate geometry.}
\label{tab:fixed_selection}\label{tab:repair_behavior}\label{tab:repair_ablation}\label{tab:retention}
\small
\setlength{\tabcolsep}{4pt}
\begin{tabular}{@{}lrrr@{}}
\toprule
Scoring evidence & AP70 $\uparrow$ & Kept $\uparrow$ & False $\downarrow$ \\
\midrule
Original scores & 25.70 & \textbf{230.00} & 447.00 \\
Box attributes & 26.56$\pm$0.09 & 225.00 & 330.00 \\
+ Receiver points & 28.32$\pm$0.08 & 227.00 & 283.33 \\
+ Both sources & \textbf{29.27}$\pm$0.08 & 228.67 & \textbf{261.67} \\
\bottomrule
\end{tabular}
\end{table}

\subsection{Experimental Settings}
\label{sec:experimental_settings}
The CoInfra~\cite{ning2025coinfra} development split contains 256 adverse-weather views at 64 times (662 car observations) and 64 clear views at 16 times (264 observations). We evaluate controlled repair on the same 320 views before and after exterior removal. Repair training uses 384 views at 96 other times.

We fix detectors and splits across three repair fits unless indicated. V2X-ViT~\cite{xu2022v2xvit} and AttFuse share CoInfra adaptation data (1,893 times) and repair budgets. Table~\ref{tab:detection} includes receiver-only detection, late fusion, five-class fine-tuning, LiDAR RCP-Mix~\cite{du2025rcpbench}, and Robo3D-DIT~\cite{kong2023robo3d}. Public benchmarks use released HEAL~\cite{lu2024heal}, FadeLead~\cite{wu2026fadelead}, and V2X-DGW~\cite{li2025v2xdgw} weights (Sections~\ref{supp:training_references} and~\ref{supp:public_detectors}).

On V2X-R, Figure~\ref{fig:problem_analysis}(b) uses 72 full-weather times. Regional interventions and point filtering use all 3,166 test times. CoAlign and both AttFuse checkpoints~\cite{huang2025v2xr,v2xr_model_release} use filters fitted on official TRAIN labels (Section~\ref{supp:simulation}).

All natural CoInfra comparisons share a support filter using both clouds and a background map. Controlled comparisons use native post-processing. AP50 and AP70 use rotated BEV IoU thresholds of 0.5 and 0.7. Table~\ref{tab:detection} specifies the AP evaluator. CoInfra repair targets Car, and mAP averages four classes.

\subsection{Detection Results}
AWCR raises V2X-ViT car AP70 by 4.22 points on the four adverse-weather CoInfra development recordings (Table~\ref{tab:detection}(a)). With the three fits fixed, the mean gain ranges from +3.69 to +5.10 AP70 points when we omit each recording in turn and pool the remaining three. November~27 was held out from detector and repair training and model selection. Across all 581 synchronized times (2,324 receiving views) in slice~6 of that heavy-snow recording, the same three fits at $\lambda=0.7$ gain 3.71 points on average (Table~\ref{tab:supp_coinfra_recordings}). On the development set, four-class mAP70 gains 1.05 points (Table~\ref{tab:supp_awcr_repeated}), and car AP70 also improves for AttFuse and Robo3D-DIT-adapted V2X-ViT.

\begin{samepage}
On the full synchronous V2X-Radar-C validation set, HEAL with AWCR reaches $68.22\pm0.04$ vehicle AP70 at $\lambda=0.7$ from CoInfra. Selecting $\lambda$ on 48 held-out times from HEAL's training recordings gives $\lambda=0$ and 67.90 AP70. A larger generic region-based refiner on the same frozen HEAL reaches 69.96 with direct quality scoring (Table~\ref{tab:supp_public_benchmarks}). FadeLead gains 0.71 points. For identical V2X-DGW predictions, AP70 rises under global ranking but falls under released frame order in all three weather conditions (Section~\ref{supp:native_dgw}).\par
\end{samepage}

In a separate V2X-R experiment, we train one peer filter per weather using simulator point labels and select it by the known test weather. Filtering improves vehicle AP70 in all six checkpoint--weather pairs (Table~\ref{tab:detection}(b)). For the same fog-trained AttFuse checkpoint in simulated fog, AP70 rises from 61.49 without filtering to $68.34\pm0.04$ with peer filtering. Fog filtering retains 99.01\% of fog-trained AttFuse's originally correct fog detections, reduces false detections by 19.1\%, and preserves clean AP70.

\subsection{Ablation and Repair Analysis}
\label{sec:component_roles}
On natural adverse-weather CoInfra inputs, car AP70 is 25.70 with the original detector, 29.27 with both-source reranking of the original boxes, and 29.92 with full AWCR (Tables~\ref{tab:detection}(a) and~\ref{tab:fixed_selection}). With box coordinates fixed ($\lambda=0$), box attributes alone yield 26.56. Receiver point statistics raise AP70 to 28.32, and adding peer point statistics raises it by another 0.95 points to 29.27. Both-source reranking recovers 7, 7, and 8 of the 108 misses whose accurate boxes scored below threshold, and 2, 1, and 1 of the nine whose above-threshold accurate boxes were suppressed (Sections~\ref{sec:candidate_diagnosis} and~\ref{supp:fixed_selection}).

The 0.65-point gain over reranking includes changes to admission, scores, and geometry. We keep final scores fixed, revert each admitted correction to its same-anchor original box, and rerun support filtering, NMS, and matching. This rollback loses 4, 2, and 2 detections that AWCR recovered and reranking alone missed in the three fits. They span five distinct target observations whose best valid original box fell just below 0.7 BEV IoU. AWCR also loses 6--8 detections per fit that reranking retained and adds false detections (Section~\ref{supp:default_attribution}). Figure~\ref{fig:qualitative} shows recovery by coordinate correction in (a) and by reranking in (c).

We revisit the 72 induced misses from Section~\ref{sec:candidate_diagnosis}. A paired recovery requires detecting the target both before and after removal. Among the 31 with an accurate original box after removal, AWCR recovers 6, 6, and 8 across the three fits, versus 4, 4, and 3 for reranking alone.

For 40 of the other 41, the annotation lets us construct an accurate box within AWCR's bounds at an already shortlisted anchor in every fit. The learned offsets produce an accurate box for only 3, 4, and 7 targets, respectively, all represented on the shortlist. Only one target is detected after scoring and suppression, in the third fit. Reverting its coordinates at fixed scores loses the detection. The same fit misses it before removal, leaving zero paired recoveries among the 41. Of 254 originally stable observations, AWCR newly misses 3, 4, and 7 on at least one input (Table~\ref{tab:supp_controlled_bridge}).

\subsection{Transfer and Cost}
\label{sec:repair_retention}

With CoInfra-adapted V2X-ViT fixed, we fit repair and select $\lambda$ on the remaining weather conditions. Pooled AP70 rises from 25.70 to $29.71\pm0.42$ across the three held-out tests. Reranking alone trails overall but leads in rain (Section~\ref{supp:weather_transfer}).

Fitted on a separate dry recording, AWCR improves F-Cooper on both DAIR/V2X-Seq test recordings at CoInfra's $\lambda=0.7$ (Table~\ref{tab:detection}). In rain, AP70 is 32.86 before repair, 37.84 with reranking alone, 37.86 with AWCR, and 45.04 for the receiver alone (Table~\ref{tab:supp_natural_rain}).

Local Correction adds 0.47\% more parameters to V2X-ViT and 1.10\% to AttFuse. From CPU-resident inputs through decoding and filtering, their median times rise from 0.294 to 0.584\,s and from 0.515 to 1.030\,s, respectively (Section~\ref{supp:cost}). FadeLead hybrid sends an extra 13.875\,MB per input on average in uncompressed source features and point coordinates (Section~\ref{supp:public_detectors}).
\par

\begingroup
\widowpenalty=10000\relax
\section{Conclusion}
\label{sec:conclusion}
Even with object returns fixed, exterior changes could cause misses through the pre-fusion representation around an object. On CoInfra, restoring these neighborhood features recovered nearly all induced misses. Each equal-area exterior control recovered none.
Across the paired V2X-R scans, changing only object returns caused more misses in simulated fog, and changing only exterior returns caused more in simulated snow.
At inference, AWCR uses current measurements to recover some targets whose accurate original boxes scored too low or were suppressed. Bounded corrections recovered additional natural detections whose original boxes fell short of the overlap threshold.
In a separate V2X-R experiment, peer filters trained on weather-specific simulator point labels improved AP70 in all six detector--weather pairs when test weather was known.
On controlled pairs, full AWCR recovered more misses than reranking alone but also lost some originally correct detections that reranking retained.
\par
\endgroup

\begin{samepage}
\section*{LLM Usage Statement}
OpenAI ChatGPT and Codex were used for literature searches, experiment design and execution, code implementation and debugging, result checks, figure preparation, and manuscript drafting and editing. ChatGPT and DeepSeek provided reviewer-style feedback on the manuscript. The authors take responsibility for all scientific claims, evidence, code, and citations in this preprint.
\par
\end{samepage}

\begingroup
\small
\bibliographystyle{ieeenat_fullname}
\bibliography{references}
\endgroup

\microtypesetup{activate=false}
\clearpage
\setcounter{page}{1}
\renewcommand{\thepage}{S\arabic{page}}
\setcounter{section}{0}
\renewcommand{\thesection}{\Alph{section}}
\renewcommand{\theHsection}{supp.\arabic{section}}
\setcounter{table}{0}
\renewcommand{\thetable}{S\arabic{table}}
\renewcommand{\theHtable}{supp.\arabic{table}}
\setcounter{equation}{0}
\renewcommand{\theequation}{S\arabic{equation}}
\renewcommand{\theHequation}{supp.\arabic{equation}}
\setcounter{figure}{0}
\renewcommand{\thefigure}{S\arabic{figure}}
\renewcommand{\theHfigure}{supp.\arabic{figure}}
\maketitlesupplementary
\raggedbottom

\setlength{\textfloatsep}{8pt plus 2pt minus 2pt}
\setlength{\dbltextfloatsep}{8pt plus 2pt minus 2pt}
\setlength{\dblfloatsep}{6pt plus 2pt minus 2pt}
\setlength{\intextsep}{6pt plus 2pt minus 2pt}
\captionsetup{skip=4pt}
\renewcommand{\arraystretch}{.98}
\makeatletter
\setlength{\@fptop}{0pt}
\setlength{\@fpsep}{8pt}
\setlength{\@fpbot}{0pt plus 1fil}
\setlength{\@dblfptop}{0pt}
\setlength{\@dblfpsep}{8pt}
\setlength{\@dblfpbot}{0pt plus 1fil}
\makeatother

\setcounter{topnumber}{4}
\setcounter{dbltopnumber}{4}
\renewcommand{\topfraction}{.95}
\renewcommand{\dbltopfraction}{.95}
\renewcommand{\textfraction}{.05}
\noindent
Section~\ref{supp:data} details data, evaluation, and implementation. Section~\ref{supp:interventions} traces controlled losses from input changes through representations to candidate evidence. For practical repair, Section~\ref{supp:components} examines components and correction weights, Section~\ref{supp:training_references} compares detector training, and Sections~\ref{supp:public_detectors} and~\ref{supp:independent_real} test released detectors and transfer across recordings. Section~\ref{supp:simulation} covers input filtering in simulated V2X-R weather. Section~\ref{supp:controlled_target_rows} lists the repair outcome for each of the 72 controlled CoInfra misses across all three fits (Table~\ref{tab:supp_controlled_target_rows}).

In candidate repair comparisons, AWCR uses the CoInfra-selected weight $\lambda=0.7$ unless another weight is specified.
At $\lambda=0$, the route selects only original candidates. At $\lambda=1$, both original and corrected candidates have full weight.
The weather-transfer study also selects weights using only weather conditions other than the one being tested (Section~\ref{supp:weather_transfer}).
A target observation is one annotated object at one time in one receiving view, so the same object can contribute several observations.
We report BEV AP on a 100-point scale, and comparisons repeated across three component fits give their mean and sample standard deviation.

\section{Data, Evaluation, and Implementation}
\label{supp:data}
\subsection{CoInfra Inputs and Detector Fitting}
\noindent\begin{minipage}{\columnwidth}
\centering
\captionof{table}{CoInfra repair-training and development allocations. Each date row counts receiving views in both directions. The last row counts physical times. The natural and clear-input repair cohorts share 16 clear times and 32 receiving views.}
\label{tab:supp_real_split}
\small
\setlength{\tabcolsep}{4pt}
\begin{tabular}{@{}llrrr@{}}
\toprule
Date & Weather & Train & Natural & Clear repair\\
\midrule
2025-03-24 & Rain & 64 & 48 & 0\\
2025-03-27 & Clear & 64 & 32 & 32\\
2025-04-07 & Snow & 64 & 32 & 0\\
2025-02-12 & Snow & 64 & 48 & 0\\
2025-04-02 & Freezing rain & 64 & 128 & 0\\
2025-11-20 & Clear & 64 & 32 & 32\\
\midrule
Views & & 384 & 320 & 64\\
Physical times & & 96 & 80 & 32\\
\bottomrule
\end{tabular}
\end{minipage}
\par\smallskip

We use roadside pairs $(6,9)$ and $(7,8)$ across the six CoInfra recordings in Table~\ref{tab:supp_real_split}. At each physical time, we evaluate both receiving directions by treating each node in turn as the receiver and the other as its peer.
Each pair contributes 160 natural-development views. Together they contain 662 visible car observations in adverse weather and 264 in clear weather.
The mechanism diagnostic and the paired repair analysis of its 72 induced misses use this full natural-development allocation.
A separate clear-input comparison uses 268 car observations from 64 views of pair $(7,8)$.
Table~\ref{tab:supp_real_split} identifies the overlap between these evaluation cohorts. Both are disjoint in physical time from repair training, including across receiving directions.

Before training repair, we adapted V2X-ViT and AttFuse on the same 1,893 times from the rain, March clear, and April snow recordings.
We evaluated epochs 5, 10, 15, and 20 on a separate set of 752 times and selected epoch 20 by the lowest mean native detection loss.
Detector training, checkpoint selection, and development evaluation use disjoint timestamps. Of the repair-training times, 48 overlap with detector training. All repair-training times are disjoint from checkpoint selection.
The CoInfra development results therefore evaluate new times within the designated recordings.
Section~\ref{supp:coinfra_recordings} evaluates an additional CoInfra heavy-snow recording. Section~\ref{supp:natural_rain_recording} tests transfer to DAIR/V2X-Seq recordings excluded from repair fitting.
\subsection{Matching and Shared Support Filtering}
\label{supp:evaluation}

The natural CoInfra comparisons in Table~\ref{tab:detection} apply the same support filter to car predictions.
We transform the current points from each source to world coordinates. A point is foreground when it lies more than $0.4\sqrt{3}$\,m from the nearest point in that source's fixed background map.
A box passes the filter if a foreground point from either source lies within its rotated 3D bounds or if its score at filtering exceeds 0.6.
AWCR admits and scores candidates before applying this filter, and native suppression follows.
The filter uses both unmodified current source clouds even when the detector receives filtered points or only the receiving view.
Controlled comparisons instead omit this filter and use native postprocessing.
The receiver references used for the natural candidate census are specified separately in Section~\ref{supp:natural_evidence}.

The repair-fitting allocation is listed in Table~\ref{tab:supp_real_split}.
For the support filter, we build the four background maps before repair fitting, using a separate allocation of 32 March 27 clear training times per source that is disjoint from development.
We keep 0.4\,m world cells that appear in at least 16 inputs, excluding annotated training boxes expanded by 1\,m horizontally and 0.5\,m vertically.

For candidate repair, class-wise non-maximum suppression (NMS) uses at most 4,096 candidates per class and suppresses lower-ranked boxes with BEV overlap greater than 0.15.
We then match rotated BEV predictions to same-class annotations greedily in score order at overlap thresholds of 0.5 and 0.7.
For AP, we rank each class's predictions jointly across evaluation inputs and use VOC2010 all-point interpolation over the full retained curve.
Detection counts use score above 0.27 and overlap at least 0.7. Unmatched predictions above the score threshold count as false detections.
CoInfra mAP averages Car, Truck, Bicycle, and Person. Bus has no ground truth in this evaluation. Because repair affects only Car, its AP change divided by four gives the mAP change.
For controlled repair pairs, we also track whether repair preserves correctness across the two inputs.
Rescued counts previously correct detections lost after removal and restored by repair on both inputs. New counts observations originally correct on both inputs but missed by repair on either. Errors sums unrecovered induced misses and these new losses.
Input filtering with Point Classifier follows each released detector's native processing and frame-order AP accumulation, as detailed in Section~\ref{supp:simulation}.
\subsection{Processing Cost}
\label{supp:cost}\label{sec:cost}
\noindent\begin{minipage}{\columnwidth}
\centering
\captionof{table}{Median FP32 processing time on a shared RTX 4080, with and without AWCR candidate repair ($\lambda=0.7$). The last column reports the median total peak GPU memory allocated by the detector with AWCR.}
\label{tab:cost}
\small
\setlength{\tabcolsep}{4pt}
\begin{tabular}{@{}lrr@{}}
    \toprule
    Detector & Native $\to$ AWCR (ms) & \shortstack[r]{Median peak GPU\\memory with AWCR (MiB)} \\
    \midrule
    V2X-ViT & 294 $\rightarrow$ 584 & 1,450.3 \\
    AttFuse & 515 $\rightarrow$ 1,030 & 525.1 \\
    \bottomrule
  \end{tabular}
\end{minipage}
\par\smallskip

To measure the additional processing required by candidate repair, we run the detector alone and with AWCR on 12 fixed real inputs.
Each method uses two warm-ups and three interleaved repeats.
Timing starts from retained inputs in CPU memory and ends after decoding all five classes and filtering the final detections.
The interval includes detector inference and repair, while disk I/O, raw preprocessing, network transfer, loading, and compilation are excluded.
Both quality scorers remain resident during timing, and the compiled point-statistics implementation preserves outputs.
Candidate repair roughly doubles median processing time for both detectors (Table~\ref{tab:cost}).

For communication, the median peer payload is 0.41\,MiB of FP32 XYZI data, excluding transport overhead.

\subsection{Local Correction}
\label{supp:local_training}

To give Local Correction comparable evidence from both sensors, we pool their features and point counts in the receiving frame. We adaptively average each source's 64-channel scatter output onto the detector's output grid.
Within each $5\times5$ neighborhood, we average features over occupied cells by dividing the mean support-weighted features by the mean occupied-cell support. We also sum the retained-point counts in that neighborhood.
For retained-point count $n_i$, support is $M_i=\mathbf{1}[n_i>0]$ and normalized density is $\rho_i=\min\{2,\log(1+n_i)/\log513\}$.
We normalize source and fused features using per-channel RMS scales from the original 384 training views, floor the scales at 0.001, and clip the normalized values to $[-16,16]$.

We concatenate four copies of the 64 source channels with density to form the 257-channel input to the shared $257\to64\to64$ encoder. It uses $1\times1$ convolutions with GELU and a scalar sigmoid gate.
A $1\times1$ GELU encoder projects the fused context to 64 channels.
The $194\to128\to16$ prediction head combines the two source encodings, fused context, and support flags. At each grid location, it outputs residuals for the two repaired-class anchor logits and 14 encoded regression coordinates.
V2X-ViT uses a $144\times144$ grid and 256 context channels, and its correction head adds 64,209 parameters (0.47\% of the detector).
AttFuse uses a $288\times288$ grid and 384 context channels, and its head adds 72,401 parameters (1.10\%).
For anchor height $h_a$, width $w_a$, length $l_a$, and BEV diagonal $d_a=\sqrt{w_a^2+l_a^2}$, the regression bounds are
\begin{equation}
\begin{aligned}
\beta_a={}&(0.5/d_a,0.5/d_a,0.25/h_a,\\
          &\phantom{(}0.12,0.12,0.12,0.2).
\end{aligned}
\label{eq:reg_bounds}
\end{equation}

These bounds follow the encoded coordinates $(x,y,z,h,w,l,\theta)$ for center offsets, log-size changes, and yaw. Lengths use metres and yaw uses radians.
Concatenating the bounds of both car anchors gives the 14-channel $\beta$ in Eq.~\eqref{eq:bounded_residual}.
The head bounds score-logit changes by three and applies zero correction to other classes.

To construct the training inputs, we use each view in its original condition and create versions of clear views with exterior points removed from the receiver, peer, or both.
Each original view and each modified version is a separate training condition. All modified versions preserve object measurements.
All conditions from the same view share six patch centers. Four centers follow a random ordering of cars, cycling when needed, and two are sampled uniformly.
For views without cars, all centers are sampled uniformly. Each $16\times16$ patch stays within the feature grid and contains its selected center. The 768 training conditions yield 4,608 patches.

We first assign positive training anchors using the visible annotations. An anchor is positive if it has BEV IoU of at least 0.6 with a same-class annotation or is a target's best anchor.
When targets compete for an anchor, the larger overlap resolves the conflict. A target without an assigned anchor takes its best unused same-class anchor.
For an exterior-loss variant, correctly matched native detections on the corresponding clear input supply additional positive anchors with annotation-derived targets.
We inspect decoded boxes at positive anchors and at anchors whose scores can exceed 0.27 with a logit increase of at most three.
A valid inspected box with BEV IoU at least 0.7 to a visible same-class annotation becomes positive and uses its current native regression as the target.

Before computing the classification loss, we check whether inspected anchors are ambiguous or cannot reach a correct detection within the allowed update.
Inspected negatives with decoded BEV IoU in $[0.5,0.7)$ have zero classification weight.
For each positive, we clip the residual toward its target coordinatewise to the correction bounds.
After applying this clipped update, we set its classification weight to zero if the resulting box has maximum BEV IoU below 0.7 or the maximum allowed logit yields a score of at most 0.27.
These checks affect classification weights only, and the box loss uses all positive anchors.

Within each patch, we normalize the classification term in Eq.~\eqref{eq:local_loss} by the sum of classification weights over positive anchors and the box term by the number of positive anchors. Each denominator is at least one.
Focal classification uses $\alpha=0.25$ and $\gamma=2$. The box term uses a Smooth L1 transition of $1/9$ and a sine yaw error.
At native-correct anchors with score $>0.27$ and overlap $\geq0.7$, the retention term averages a squared confidence-drop hinge with margin 0.05 and mean squared regression drift normalized by $\beta$.

We initialize the final layer to zero so the head starts from the original predictions.
We fit with AdamW for 3,000 updates of four patches and evaluate the final update.
The learning rate starts at $3\times10^{-4}$ and decays to $3\times10^{-5}$ on a cosine schedule. Weight decay is $10^{-4}$ and gradients are clipped at norm five.
The fixed run uses seed 20261019, and Section~\ref{supp:seed_stability} describes repeated fits.
\subsection{Quality Selection}
\label{supp:quality_training}

Quality Selection describes each candidate through its box attributes and the current point measurements from each source.
The seven box attributes are the confidence logit, height, width, length, sine and cosine of heading, and a branch indicator that is zero for original candidates and one for corrected candidates.
Each source contributes 30 point statistics in rotated box-centered coordinates $(\xi,\eta,\zeta)$, normalized by box dimensions with divisor floors of 0.1\,m.
The support region uses $|\xi|,|\eta|<0.6$ and $|\zeta|<1$. Within it, points with $|\xi|,|\eta|<0.5$ are interior points.
We record two log-counts for the interior and nearby exterior, 16 interior log-counts from a $4\times4$ histogram, means and standard deviations of $(\xi,\eta,\zeta)$, and extrema of $\xi$ and $\eta$.
The remaining two statistics are the fractions satisfying $|\xi|>0.4$ and $|\eta|>0.4$, respectively. A source with empty support contributes zeros.

We fit the quality regressors with histogram gradient boosting and squared error.
For each training candidate, the regression target is its maximum BEV overlap with a visible annotated car in the current input, or zero if no car is present. Each date contributes equal total training weight.
Fitting uses 300 iterations, learning rate 0.06, 15 leaves, at least 40 samples per leaf, an $L_2$ penalty of one, and no early stopping.
To fit $R_1$ on both branches, we reuse the six patches per condition and rank anchors by the larger of their original and corrected confidence.
The highest-ranked 512 anchors and up to 32 deterministically sampled from the remainder supply candidates from both branches.
We fit $R_0$ separately on original candidates with branch indicator zero.
For this original-candidate regressor, we deduplicate anchors across patches and use the same sampling rule of the top 512 anchors plus up to 32 others. This gives 402,139 training candidates for V2X-ViT and 374,286 for AttFuse.
Both regressors use the same target definition, date weighting, 67 features, and boosting settings.

To compare the point evidence used at full correction weight, the variant without point statistics zeros all 60 entries. Variants using only receiver or peer statistics zero the other source's 30 entries. These variants share the same 814,285 training candidates, targets, order, and weights.

At inference, we clip the unweighted confidence to $[10^{-6},1-10^{-6}]$ for the descriptor. For rescoring, we clip it to $[10^{-9},1-10^{-9}]$ and clip rescored logits to $[-40,40]$.
After the weighted admission of up to 4,096 candidates, Eq.~\eqref{eq:correction_trust} blends $Q_0$ and $Q_1$ on original boxes and scales corrected scores. Anchor identifiers break score ties deterministically.

\subsection{Correction-Weight Selection}
\label{supp:correction_trust_selection}
We select the default correction weight within the CoInfra repair-training allocation.
We divide it into 60 fitting, 12 buffer, and 24 validation times, corresponding to 240, 48, and 96 views.
We train a Local Correction head and both quality regressors on the fitting subset. The detector checkpoint stays fixed, and feature normalization reuses the scales from the 384 repair-training views.
We evaluate Car AP70 on original validation inputs for $\lambda\in\{0,0.1,\ldots,1\}$ and choose the smallest weight within 0.1 AP points of the maximum.
This rule fixes the default at $\lambda=0.7$ before evaluation.

The main all-weather comparison and repeated fits use this default, which we also transfer to AttFuse and DAIR.
The same rule can select a weight for another model or dataset using validation inputs from its training allocation.
In the weather-transfer study, we train repair and select the weight using only the remaining weather conditions.
Comparisons at $\lambda=0$ and $\lambda=1$ use the stated weights.

\section{Controlled Diagnosis and Candidate Evidence}
\label{supp:interventions}

\begin{table*}[!tbp]
\centering
\begin{minipage}{\textwidth}
\centering
\captionof{table}{Representation restoration on all 320 natural-development views with 926 car observations. (a) Each region entry averages across views. Masks have equal area within each view and average 85.82 cells. Receiver and Peer give current occupancy fractions, Radius gives mean distance from the receiver, and RMS gives root-mean-square replacement magnitude across both sources and 256 content channels. (b) Recovery follows the 72 induced losses and 254 stable observations. Help counts restored detections among the 47 cooperative benefits lost under deletion. Gained and Lost report changes in correctness across all target observations relative to the changed input. Restoration uses original features and annotated regions for diagnosis.}
\label{tab:supp_restoration_masks}\label{tab:supp_real_restoration}
\small
\setlength{\tabcolsep}{4pt}
(a) Region controls\par\smallskip
\begin{tabular}{@{}lrrrr@{}}
    \toprule
    Region & Radius (m) & Receiver & Peer & RMS \\
    \midrule
    Object & 43.02 & 0.177 & 0.176 & 0.0343 \\
    Exterior $90^\circ$ & 43.16 & 0.073 & 0.146 & 0.0198 \\
    Exterior $180^\circ$ & 43.12 & 0.169 & 0.061 & 0.0316 \\
    Exterior $270^\circ$ & 43.16 & 0.063 & 0.078 & 0.0256 \\
    \bottomrule
  \end{tabular}

\par\smallskip
(b) Full-set effects and recovery\par\smallskip
\begin{tabular}{@{}lrrrrrrrrr@{}}
    \toprule
    Input / restored region & Rescued & New losses & Errors & Help & Gained & Lost & Correct & False & AP@0.7 \\
    \midrule
    Original input & 72 & 0 & 0 & 47 & 72 & 18 & 326 & 855 & 24.756 \\
    Changed input & 0 & 0 & 72 & 0 & 0 & 0 & 272 & 1012 & 19.322 \\
    Object & 71 & 0 & 1 & 46 & 73 & 18 & 327 & 962 & 24.815 \\
    Exterior $90^\circ$ & 0 & 0 & 72 & 0 & 0 & 0 & 272 & 1011 & 19.308 \\
    Exterior $180^\circ$ & 0 & 0 & 72 & 0 & 0 & 0 & 272 & 1011 & 19.342 \\
    Exterior $270^\circ$ & 0 & 0 & 72 & 0 & 0 & 0 & 272 & 1013 & 19.320 \\
    \bottomrule
  \end{tabular}
\end{minipage}
\par\smallskip

\end{table*}

\subsection{Exterior-Loss Construction}

To construct exterior-removal inputs, we identify directions where adverse-weather recordings have persistently fewer returns than clear recordings. The CoInfra profiles mark these directions in sensor-local $2^\circ\times2^\circ$ angular cells.
We compute the profiles from point counts at 16 fixed times per recording, using 16 March 27 clear training times as the reference. All these times are disjoint from development.
For cells with clear median count $r\geq2$, we mark a cell when its adverse-weather median falls below $r/4$ and at least 12 of the 16 adverse frames also fall below $r/4$.
We use the previously computed profiles for nodes 7 and 8. For nodes 6 and 9, we apply the same rule to the 16 predeclared March 27 clear and 16 April 2 freezing-rain training times.

We apply the fixed freezing-rain profile to the natural-development and clear-input repair cohorts.
Deletion protects all points inside annotated 3D boxes. We compact the surviving XYZ and intensity rows in their original order, update cell counts, and leave the discarded points unreplaced.

For simulation, we use the fixed freezing-rain profiles from CoInfra nodes 7 and 8.
Their deletion fractions, 0.38049 and 0.31667, are the median fractions of exterior points removed on the real clear-input repair cohort.
The simulation test uses periodic azimuth $[-180^\circ,180^\circ]$ and maps elevations from $[-25^\circ,2^\circ]$ to $[-16^\circ,16^\circ]$ with clipped ends.
For each simulated source, we apply the fraction to its retained original exterior points, protecting object interiors and generated weather returns.
We remove the eligible points nearest the marked cell centers to reach this count.
The receiver uses the node 8 profile, while odd- and even-indexed peers use nodes 7 and 8, respectively.

Random removal matches the directional mask's deletion count for each source. The full mask induces 31 V2X-ViT misses on the 64 clear CoInfra views (Table~\ref{tab:supp_full_scope_diagnosis}).
At 24 clear V2X-R times, applying up to 75\% of the mapped mask induces at most one V2X-ViT miss. The full mask loses 8 of 216 originally correct observations and gains 3. Equal-count random removal loses 2--4 and gains none across three seeds (20261024--20261026). AttFuse loses 3 of 219 and gains 4 under the full mask, versus 2 lost and 1 gained under random removal (one seed).

\subsection{Weather Changes Inside and Outside Objects}
\label{supp:weather_counterfactual}

For each of the 3,166 times in the 16 official V2X-R test scenes, we use the paired clear, fog (\texttt{fog\_0.060}), and snow (\texttt{snow\_2.5\_2.0}) scans. The fixed V2X-ViT detector uses the released 70\,m communication range. A peer is present at 3,133 times, and only the receiver is present at the other 33.
For each source, we separate points inside any annotated vehicle box from the exterior in receiver coordinates. Replacing either region of the clear scan with the corresponding weather returns gives two single-region interventions. We also evaluate the clear scan and the scan with weather in both regions. All four inputs share the clear-reference labels and diagnostic postprocessing. The annotations serve only to construct the intervention and score detections.

\begin{table*}[!tbp]
\centering
\caption{Paired weather interventions by V2X-R test scene. Clear correct counts correctly detected observations among 26,777 vehicle observations shared by both weather tests. Object and Exterior count clear detections lost when only that region changes. Recovered counts exterior-induced misses found after restoring clear pre-fusion features in all annotated object neighborhoods. A vehicle can contribute at multiple times. Scene IDs give the 2024 recording date (month/day) and start time.}
\label{tab:supp_weather_counterfactual}
\small
\setlength{\tabcolsep}{5pt}
\begin{tabular}{@{}lrr rrr rrr@{}}
\toprule
& & & \multicolumn{3}{c}{Fog} & \multicolumn{3}{c}{Snow} \\
\cmidrule(lr){4-6}\cmidrule(l){7-9}
Scene ID & Times & Clear correct & Object & Exterior & Recovered & Object & Exterior & Recovered \\
\midrule
06/27 16:27:32 & 376 & 4,959 & 2,461 & 1,069 & 1,050 & 878 & 1,863 & 1,843 \\
06/27 17:05:43 & 214 & 2,560 & 1,183 & 516 & 507 & 560 & 793 & 788 \\
06/27 17:41:33 & 332 & 1,686 & 748 & 406 & 393 & 256 & 874 & 859 \\
06/27 19:09:42 & 278 & 2,852 & 359 & 1,278 & 1,258 & 282 & 2,513 & 2,495 \\
06/28 09:17:57 & 178 & 710 & 170 & 168 & 164 & 58 & 384 & 383 \\
06/28 09:33:41 & 75 & 641 & 167 & 128 & 125 & 39 & 293 & 291 \\
06/28 09:52:03 & 79 & 832 & 542 & 154 & 142 & 250 & 182 & 179 \\
06/28 10:07:17 & 111 & 45 & 10 & 14 & 14 & 7 & 41 & 40 \\
06/28 10:22:46 & 211 & 2,383 & 1,441 & 272 & 228 & 780 & 539 & 508 \\
06/28 10:58:01 & 88 & 249 & 135 & 29 & 21 & 54 & 79 & 73 \\
06/28 11:17:03 & 114 & 1,211 & 830 & 133 & 112 & 494 & 254 & 238 \\
06/28 14:59:30 & 173 & 1,504 & 511 & 591 & 580 & 313 & 808 & 800 \\
06/28 15:26:15 & 149 & 339 & 137 & 171 & 165 & 87 & 219 & 217 \\
06/28 15:52:12 & 322 & 817 & 222 & 434 & 422 & 151 & 512 & 510 \\
06/28 16:18:18 & 130 & 492 & 189 & 146 & 145 & 128 & 191 & 191 \\
06/28 16:36:53 & 336 & 1,362 & 598 & 265 & 256 & 402 & 547 & 535 \\
\midrule
Total & 3,166 & 22,642 & 9,703 & 5,774 & 5,582 & 4,739 & 10,092 & 9,950 \\
\bottomrule
\end{tabular}
\end{table*}

Exterior-only changes induce losses in every scene. They affect 232 distinct scene--track pairs across 2,519 times in fog and 243 across 2,830 times in snow. Object-only changes cause more losses than exterior-only changes in 11 of 16 fog scenes. The exterior causes more losses in 13 of 16 snow scenes (Table~\ref{tab:supp_weather_counterfactual}).
Changing both regions gives 14,717 losses in fog and 14,889 in snow. Of these, 1,733 and 1,992, respectively, occur only when both regions change. Conversely, 64 fog and 93 snow observations lost under exterior-only weather are detected when both regions change. The single-region counts therefore cannot be added to assign contributions to complete-weather loss.

On each exterior-only input, we restore the clear-reference pre-fusion content features in the union of annotated object neighborhoods. We project the boxes onto the model grid and expand them by half a cell diagonal (Section~\ref{supp:restoration}). Each exterior control rotates the object mask through the indicated angle, excludes object cells, and matches the original cell count.
Restricting the comparison to the 3,133 times with an active peer yields 5,545/5,736 recoveries in fog and 9,882/10,024 in snow. The highest equal-area exterior counts are 24 and 14, respectively.

\begin{table*}[!tbp]
\centering
\caption{Full-test feature restoration after exterior-only weather. The denominators are exterior-induced misses, clear detections that survived the change, and detections gained through the change, respectively. The last two columns count matched vehicle observations and unmatched predictions after restoration.}
\label{tab:supp_weather_restoration}
\small
\setlength{\tabcolsep}{6pt}
\begin{tabular}{@{}llrrrrr@{}}
\toprule
Weather & Restored region & Recovered & New losses & Reversed gains & Correct & False \\
\midrule
Fog & Object & 5,582/5,774 & 102/16,868 & 332/402 & 22,491 & 2,493 \\
& Exterior $90^\circ$ & 16/5,774 & 5/16,868 & 2/402 & 17,279 & 2,470 \\
& Exterior $180^\circ$ & 24/5,774 & 9/16,868 & 1/402 & 17,284 & 2,459 \\
& Exterior $270^\circ$ & 6/5,774 & 3/16,868 & 0/402 & 17,274 & 2,448 \\
\midrule
Snow & Object & 9,950/10,092 & 37/12,550 & 242/297 & 22,623 & 3,253 \\
& Exterior $90^\circ$ & 14/10,092 & 6/12,550 & 0/297 & 12,856 & 3,171 \\
& Exterior $180^\circ$ & 14/10,092 & 24/12,550 & 2/297 & 12,837 & 3,167 \\
& Exterior $270^\circ$ & 8/10,092 & 1/12,550 & 0/297 & 12,854 & 3,169 \\
\bottomrule
\end{tabular}
\end{table*}

Object-neighborhood restoration recovers more induced misses than any exterior control in every scene. It also reverses some detections gained through the exterior change and introduces new losses (Table~\ref{tab:supp_weather_restoration}). These restorations use clear-reference features for diagnosis. Candidate repair uses current observations (Section~\ref{supp:controlled_candidates}).

For all inputs we place object-region point rows first, preserving coordinates, intensity, and order within each region. After cell sampling, the retained object rows match between clear and exterior-only inputs, and between object-only and complete-weather inputs. At the original 24 times from three scenes, the full-test run reproduces the archived target decisions and matched boxes. Among 216 clear detections in that subset, the object-only and exterior-only losses are 131 and 32 in fog, and 61 and 57 in snow. On this subset, using native point order reproduces the original dense outputs to absolute and relative tolerance $10^{-5}$. The common order leaves clear and complete-weather hit decisions unchanged. It shifts globally ranked AP70 by $-0.01$ points on clear inputs, $+0.61$ in complete fog, and $-0.01$ in complete snow, with one additional false detection in fog.

Fixed object returns can still share an input cell with changed exterior returns. On the same 24-time subset, 1,966 of 13,768 target source-cell encodings change in fog and 1,478 in snow. Section~\ref{supp:encoding_control} protects entire occupied target cells to test propagation beyond this encoding step.

\subsection{Controlled Diagnosis across Public Recordings}
\label{supp:public_mechanism}

We apply the same controlled deletion test with the released LiDAR F-Cooper detector~\cite{chen2019fcooper} on three V2X-Radar-C validation recordings~\cite{yang2025v2xradar}.
Recordings 1--3 are \texttt{2024-05-13-15-23-14-B}, \texttt{2024-05-15-16-45-02-C}, and \texttt{2024-05-07-18-42-36}.
Sampling every sixth frame gives 32, 32, and 20 times, respectively, and 168 views across both receiving directions.
Because these recordings lack weather labels, we use them to test the response to controlled exterior removal.
The released checkpoint's full pretraining provenance is unverified.

We use F-Cooper's native preprocessing, field of view, coordinates, decoding, $128\times128$ output grid, and two anchors.
The Vehicle category groups cars, buses, and trucks.
Matching and NMS use the common settings, and the background-map filter applies only to CoInfra.
The vehicle and roadside inputs use the two fixed CoInfra exterior-loss profiles and an equal-count random control with seed 20261024.
Both variants protect object interiors and preserve the values and order of surviving points.

Table~\ref{tab:mechanism} compares directional and random losses across these recordings.
Among the directional losses, 12, 103, and 7 detections, respectively, were cooperative benefits before removal.

For restoration, we take the two aligned $256$-channel feature maps immediately before F-Cooper's max fusion. On the $128\times128$ grid (1.6\,m cells), we project both sources' vehicle, pedestrian, and cyclist boxes into the receiver frame. The object mask combines their BEV footprints, each expanded along both box axes by half a cell diagonal. Each exterior control rotates this mask around the receiver and replaces overlapping cells to match its area (Section~\ref{supp:restoration}). Area matching allows occupancy and the magnitude of feature changes to differ.
We restore all channels from the pre-removal input inside the selected mask on both sources, then run native fusion, detection, and NMS. The denominator for recovery is the set of originally correct Vehicle observations missed after removal. These observations span 14, 29, and 4 distinct annotation IDs in recordings 1--3.
Table~\ref{tab:supp_radarc_restoration} also tracks outcomes outside that denominator. Object restoration loses 19 of 1,706 detections that survived removal and reverses 75 of the 115 detections gained under removal.

\begin{table}[!tbp]
\centering
\caption{Pre-fusion feature restoration on the three V2X-Radar-C recordings. Lost counts removal-induced misses. Object and Exterior give recoveries from those misses, with exterior counts ordered by rotation ($90^\circ$, $180^\circ$, $270^\circ$). New gives losses among detections correct both before and after removal. Reversed gives losses among detections gained under removal. The last two columns use object restoration.}
\label{tab:supp_radarc_restoration}
\small
\setlength{\tabcolsep}{3pt}
\begin{tabular}{@{}lrrrrr@{}}
\toprule
Rec. & Lost & Object & Exterior & New & Reversed \\
\midrule
1 & 23 & 16 & 1/0/2 & 7/723 & 20/28 \\
2 & 252 & 187 & 8/2/10 & 8/815 & 42/69 \\
3 & 12 & 11 & 2/0/3 & 4/168 & 13/18 \\
\midrule
All & 287 & 214 & 11/2/15 & 19/1,706 & 75/115 \\
\bottomrule
\end{tabular}
\end{table}

\subsection{Original-Feature Restoration}
\label{supp:restoration}

The deletion test identifies detections that are sensitive to exterior changes. To test the role of object-neighborhood features in those misses, we restore their values from the same input before deletion under its observed weather.
The real V2X-ViT fusion input has shape $1\times2\times144\times144\times259$ with 1.6\,m spatial cells.
We replace its first 256 content channels within the selected region and continue the detector with the three prior channels and subsequent computation intact.
The object region is the union of every annotated box projected into BEV and expanded along each axis by half the output-cell diagonal. We apply the same mask to both sources.

The exterior controls match the object region's cell count.
For each control, we rotate the object mask by $90^\circ$, $180^\circ$, or $270^\circ$, remove any object overlap, and fill the region in order of distance to the rotated mask, radial proximity, and cell index.
This construction gives regions with similar mean range, although their occupancy and replacement magnitude differ (Table~\ref{tab:supp_real_restoration}).

Recovery counts track previously correct observations that deletion turns into misses. Gained and Lost compare all target observations with the modified input, including targets that become detectable only after deletion.
Table~\ref{tab:supp_full_scope_diagnosis} separates the results by observed weather and roadside pair and identifies the clear-input repair cohort used below.

At the 24 clear V2X-R times tested with the mapped angular mask, object-neighborhood restoration recovers seven of the eight V2X-ViT losses, while replacing the exterior complement recovers four.
Restoring the entire receiver or all peers recovers five each.
Each of these interventions preserves the 208 originally stable observations.
The exterior complement in this earlier comparison covers the full region outside the object mask, whereas the weather controls in Section~\ref{supp:weather_counterfactual} use three equal-area regions.

\begin{table*}[!tbp]
\centering
\small
\caption{Controlled exterior-removal diagnosis with fixed V2X-ViT. GT counts visible car observations, and Original counts correct observations before deletion. Object and Exterior count recovered induced losses. Exterior gives the count for each of the three controls. New/stable gives object-restoration losses among previously stable observations. Weather and pair rows stratify the full natural cohort. The clear-input repair cohort shares 32 views at 16 times with that cohort.}
\label{tab:supp_full_scope_diagnosis}
\setlength{\tabcolsep}{5pt}
\begin{tabular}{@{}lrrrrrrr@{}}
\toprule
Cohort & Views/times & GT & Original & Lost & Object & Exterior & New/stable \\
\midrule
All natural & 320/80 & 926 & 326 & 72 & 71 & 0 & 0/254 \\
Natural clear & 64/16 & 264 & 96 & 21 & 21 & 0 & 0/75 \\
Natural adverse & 256/64 & 662 & 230 & 51 & 50 & 0 & 0/179 \\
Natural pair $(7,8)$ & 160/80 & 504 & 272 & 65 & 64 & 0 & 0/207 \\
Natural pair $(6,9)$ & 160/80 & 422 & 54 & 7 & 7 & 0 & 0/47 \\
\midrule
Clear-input repair & 64/32 & 268 & 150 & 31 & 30 & 0 & 0/119 \\
\bottomrule
\end{tabular}
\end{table*}

\subsection{Receiver and Cooperative Responses}
\label{supp:receiver_cooperation}
On the same 320 CoInfra views, the fixed V2X-ViT uses its native post-processing for both modes. After exterior removal, the receiver alone loses 30 of 155 originally correct observations and gains 14. Using both altered clouds, cooperation loses 72 of 326 and gains 18. Among the 90 adverse-weather observations originally detected by both modes, each loses 14. Ten losses are shared and four are unique to each mode. In 47 of the 72 cooperative losses, only cooperation had detected the target before removal, and the receiver alone still misses it afterward.

\subsection{Propagation with Fixed Target-Cell Encodings}
\label{supp:encoding_control}

Preserving points inside a box need not preserve their input encoding because deleted exterior returns can share the same input cell.
We therefore repeat the CoInfra diagnostic while protecting every complete stored input cell containing a point inside any annotated box.
Only points outside these protected cells remain eligible for the original directional mask.
Across all 320 views, the point rows, counts, and encoded features of 28,062 protected source-cell entries are exactly unchanged in the 640 source inputs.

Protecting the complete cells slightly reduces the deletion count.
To separate this change in count from the choice of protected cells, we also sample a fixed subset of the original directional mask with exactly the protected condition's deletion count in each source, using seed 20261024.
The subset adds no deletion outside the original mask.
Table~\ref{tab:supp_encoding_control} reports all gained and lost detections under the unchanged diagnostic post-processing.
The original and protected masks remove 32.50\% and 32.30\% of input points, respectively.

\begin{table*}[!tbp]
\centering
\caption{Complete input-cell protection on the full CoInfra diagnostic cohort. All rows share 926 target observations and 326 originally correct detections. Lost and Gained compare with Original. The count-matched subset removes exactly as many points per source as the protected condition. Removed is the fraction of all stored input points deleted.}
\label{tab:supp_encoding_control}
\small
\setlength{\tabcolsep}{6pt}
\begin{tabular}{@{}lrrrrr@{}}
\toprule
Input & Removed (\%) & Lost & Gained & Correct & False \\
\midrule
Original & 0.00 & 0 & 0 & 326 & 855 \\
Directional removal & 32.50 & 72 & 18 & 272 & 1,012 \\
Complete target-cell protection & 32.30 & 71 & 18 & 273 & 1,010 \\
Count-matched subset & 32.30 & 58 & 15 & 283 & 1,060 \\
\bottomrule
\end{tabular}
\end{table*}

Of the original 72 losses, 68 remain under complete cell protection and four are retained.
Three other originally correct observations become misses, giving 71 losses in total.
Thus, most of the original failures persist with identical target-cell encodings.

We next compare local feature restoration under both input conditions at four fixed network sites, starting at the scattered input encoding, followed by the first BEV block, the fusion input, and the fused representation.
At each site, annotated BEV boxes are expanded by half the stage-cell diagonal to define the object neighborhoods.
A fixed $180^\circ$ rotation provides a disjoint region with the same number of cells.
The regions can differ in occupancy and replacement magnitude.
At the scattered input site, the neighborhood mask also contains nearby cells beyond the protected occupied cells.
Restoring the entire stage reproduces the original dense outputs, and replaying unchanged features reproduces the changed outputs within absolute and relative FP32 tolerance $10^{-5}$.

\begin{table*}[!tbp]
\centering
\caption{Four-stage feature restoration with fixed V2X-ViT on the full CoInfra diagnostic cohort. Recovered follows the induced losses in each input condition, and New follows observations correct before and after deletion. The object region is the union of all annotated neighborhoods. Exterior uses the fixed equal-area $180^\circ$ control.}
\label{tab:supp_stage_restoration}
\small
\setlength{\tabcolsep}{5pt}
\begin{tabular}{@{}llrrrr@{}}
\toprule
& & \multicolumn{2}{c}{Object region} & \multicolumn{2}{c}{Exterior region} \\
Input & Restored site & Recovered & New & Recovered & New \\
\midrule
Directional removal & Scattered input encoding & 7/72 & 7/254 & 0/72 & 0/254 \\
& First BEV block & 27/72 & 2/254 & 0/72 & 0/254 \\
& Pre-fusion & 71/72 & 0/254 & 0/72 & 0/254 \\
& Fused representation & 72/72 & 0/254 & 0/72 & 0/254 \\
\midrule
Complete target-cell protection & Scattered input encoding & 6/71 & 7/255 & 0/71 & 0/255 \\
& First BEV block & 24/71 & 1/255 & 0/71 & 0/255 \\
& Pre-fusion & 70/71 & 0/255 & 0/71 & 0/255 \\
& Fused representation & 71/71 & 0/255 & 0/71 & 0/255 \\
\bottomrule
\end{tabular}
\end{table*}

To check whether joint restoration is needed for these recoveries, we repeat the first-block and pre-fusion interventions one target neighborhood at a time.
The checks cover all 72 losses under ordinary removal and all 71 under cell protection.
They reproduce the respective object-region recovery counts in Table~\ref{tab:supp_stage_restoration}, while the individual exterior regions recover none.
Later local restoration recovers more detections.
Since subsequent processing can expose an earlier replacement to changed exterior information again, these counts identify effective intervention sites rather than the first layer at which a failure originates.

\subsection{Exterior Sensitivity after Robustness Training}
\label{supp:robust_mechanism}
On the 64 clear-input repair views, the Robo3D-DIT-adapted V2X-ViT (Section~\ref{supp:training_references}) detects 141 target observations before deletion. After directional exterior removal, AP70 rises from 43.48 to 46.50, yet 19 originally correct observations become misses and 20 previously missed observations are detected. Equal-count random removal loses 8--10 and gains 4--6 across three seeds. We follow the 19 directional losses through feature restoration. Restoring the object neighborhood recovers 18 and preserves all 122 stable detections. None of the three equal-area exterior controls recovers a miss. Object restoration also reverses gains from deletion, leaving 142 correct detections as AP70 falls to 45.09.

\subsection{Scores, Geometry, and Available Candidates}
\label{supp:decision_decomposition}

The representation interventions show that exterior changes can affect detection even when the occupied target-cell encodings are fixed.
To examine how these changes reach the final detections, we cross the dense classification scores and box regressions from the original and changed inputs.
For original outputs $(s_0,b_0)$ and changed outputs $(s_1,b_1)$, the two mixed replays use $(s_0,b_1)$ and $(s_1,b_0)$.
Decoding, candidate admission, non-maximum suppression, and matching remain unchanged.

\begin{table}[!tbp]
\centering
\caption{Output-replacement recovery with fixed V2X-ViT on the full CoInfra diagnostic cohort. Original scores or coordinates are restored in the complete output grid. Rows partition the induced losses for each input condition. The mixed replays retain all subsequent candidate filtering and suppression.}
\label{tab:supp_decision_decomposition}
\small
\setlength{\tabcolsep}{4pt}
\begin{tabular}{@{}lrr@{}}
\toprule
Recovery pattern & Directional & Protected cells \\
\midrule
Only coordinates suffice & 37 & 36 \\
Only scores suffice & 16 & 18 \\
Either suffices & 0 & 2 \\
Both required & 19 & 15 \\
\midrule
Induced losses & 72 & 71 \\
\bottomrule
\end{tabular}
\end{table}

Restoring coordinates alone recovers more losses than restoring scores alone in this comparison.
Because either replacement can also change admission and suppression, Table~\ref{tab:supp_decision_decomposition} reports which replacements suffice for recovery rather than independent shares of the underlying causes.

\phantomsection\label{supp:full_candidate_evidence}
We next ask whether the same 72 lost targets in the full diagnostic cohort still have an accurate box available for selection.
For each, we inspect every geometrically valid decoded original car candidate after exterior removal.
An accurate box has BEV overlap of at least 0.7 with the target.
Thirty-one lost targets retain at least one accurate original candidate, while 41 have none in that full pool.
Only 16 retain accurate geometry at the anchor that supplied their originally matched prediction.
For the other 15 targets with accurate candidates, the available accurate geometry lies at a different anchor.
These counts describe available geometry before the 4,096-candidate limit, distinguishing targets with accurate boxes still present from those that need a new candidate location.
They also clarify why original-score replacement and Quality Selection need not recover the same targets. The former tests one particular ranking, while the latter can rank the remaining accurate candidates differently.

Among originally correct detections, directional removal induces misses in 28 of 180 observations with original confidence of at least 0.5 and in 21 of 144 with original BEV overlap of at least 0.8.

\subsection{Current Candidates in Controlled Failures}
\label{supp:controlled_candidates}

We replay the three frozen fits on all 320 views before and after deletion. A paired rescue requires a hit on both inputs. For the 254 originally stable observations, a miss on either input counts as a new loss. Rollback keeps the 4,096 admitted candidate identities and scores fixed, restores same-anchor coordinates, and reruns joint NMS and matching. Tables~\ref{tab:supp_controlled_bridge}(a) and~\ref{tab:supp_controlled_target_rows} give aggregate and per-target outcomes.

For the 41 targets lacking an accurate original box, we test all 41,472 car anchors, excluding those too distant to overlap even at maximum shift and size. We move the box toward the annotation within Eq.~\eqref{eq:reg_bounds} and check that its center lies in range. In every fit, the same 40 targets have a feasible box at an admitted correction anchor (minimum BEV IoU 0.7600). For target 1405, we subdivide the allowed center, size and yaw intervals. An enclosing rectangle bounds the intersection from above. Combining it with the target and minimum box areas bounds the union from below. Every anchor's IoU upper bound is below 0.7.

Of the 40 reachable targets, the three fits predict accurate boxes for 3, 4, and 7, respectively. All keep an accurate box on the shortlist, but only 0, 0, and 2 have one scoring above 0.27. After joint NMS, the only hit after deletion is target 134 in the third fit (view 9, March 27). Its same-anchor IoU rises from 0.6929 to 0.7033, and rollback at the fixed score of 0.3975 loses the detection. That fit misses the same observation before deletion, so none of the 41 is recovered on both inputs in any fit.

Among the other 31 misses, every fit has one paired rescue that is lost when coordinates are rolled back. All three refer to the same observation of target 141. An original candidate at another anchor already has IoU 0.8373. The anchor used by AWCR starts at 0.6933 and reaches 0.7014, 0.7106, and 0.7217 across the fits.

The 64 clear views have 31 induced misses and 119 stable observations (Table~\ref{tab:supp_full_scope_diagnosis}). They overlap the 320-view cohort at 32 views from 16 times, including target 141, whose geometric recovery counts once. Of the 31 misses, 19 retain an accurate cooperative box within the 4,096 slots and 12 lack one across all valid original boxes. Among the 21 lost cooperative benefits, the split is 12/9.

Ten of the 31 retain a correct single-source output, 12 have an accurate box only in the saved joint pool, and nine lack both. In simulation, six of eight induced misses retain a correct single-source detection and the other two retain an accurate box in the full pool.

Table~\ref{tab:supp_controlled_bridge}(b) gives the 64-view outcomes. Table~\ref{tab:supp_fixed_selection}(b) compares the evidence used to score original boxes.

\begin{table}[!tbp]
\centering
\caption{Controlled CoInfra repair with a fixed V2X-ViT detector and three frozen fits (20261019, 20261027, 20261028, in order). Reranking uses original candidates. AWCR and its fixed-score coordinate rollback use $\lambda=0.7$. Panel (a) groups the 72 misses by availability of an accurate original box after deletion and tracks new losses among the 254 stable observations. Panel (b) gives paired rescues and new losses on 64 clear-input views.}
\label{tab:supp_controlled_bridge}
\small
\setlength{\tabcolsep}{2.6pt}
(a) Full diagnostic cohort (320 views)\par\smallskip
\begin{tabular}{@{}lrrr@{}}
\toprule
Outcome & Rerank & AWCR & Rollback \\
\midrule
Paired rescues, accurate box (31) & 4/4/3 & 6/6/8 & 5/5/7 \\
Paired rescues, no accurate box (41) & 0/0/0 & 0/0/0 & 0/0/0 \\
Hits after deletion, same 41 & 0/0/0 & 0/0/1 & 0/0/0 \\
New losses on either input (254) & 2/3/3 & 3/4/7 & 3/3/6 \\
New losses after deletion (254) & 1/2/1 & 2/3/5 & 2/2/4 \\
\bottomrule
\end{tabular}
\par\smallskip
(b) Clear-input cohort (64 views, 31 misses, 119 stable)\par\smallskip
\begin{tabular}{@{}lrrr@{}}
\toprule
Configuration & Fit 1 & Fit 2 & Fit 3 \\
\midrule
Original-candidate reranking & 4 / 0 & 5 / 0 & 5 / 0 \\
AWCR & 8 / 1 & 6 / 1 & 8 / 2 \\
Fixed-score rollback & 7 / 1 & 5 / 2 & 7 / 2 \\
AWCR w/o Point Stats & 8 / 5 & 7 / 4 & 7 / 4 \\
\bottomrule
\end{tabular}
\end{table}

Every paired rescue on the 64 views retains an accurate original in the shortlist after deletion. We fix candidate identities and correction outputs and zero all point statistics when fitting and testing the quality regressors. Mean rescues stay unchanged. With point statistics, AWCR keeps all stable observations retained without them and 4, 3, and 2 more across fits. Without point statistics, all accurate boxes score below threshold after deletion in 4, 1, and 2 of these cases. The other two are lost on the original clear inputs. Section~\ref{supp:default_attribution} examines geometric recovery on natural inputs.

\subsection{Available Evidence in Natural Misses}
\label{supp:natural_evidence}

On the unmodified natural-development inputs, we use saved V2X-ViT outputs to check which misses retain accurate geometry.
We inspect every valid decoded car anchor before the 4,096-candidate limit. An accurate candidate has BEV overlap of at least 0.7 with the target, and final detections must exceed score 0.27.
We use annotations to identify accurate candidates retrospectively.

The receiver reference uses archived V2X-ViT detections before common support filtering.
For the cooperative outputs, all 926 hit decisions agree with and without that filter.
In the AttFuse comparison, we apply shared two-source support filtering to both cooperative and receiver-only predictions. AttFuse misses 12 adverse-weather and five clear observations that remain correctly detected from the receiving view.
The intermediate-stage census uses V2X-ViT's saved candidate pools.

Table~\ref{tab:supp_natural_evidence} traces misses through the detection stages.
For most adverse-weather misses with accurate cooperative candidates, all such candidates fall below the score threshold. In the remaining cases, NMS suppresses an above-threshold accurate candidate.
The candidate limit and shared support filter account for zero additional misses.
Misses with an accurate cooperative candidate or a correct receiver detection span 49 physical times and both roadside pairs.
The table also counts cooperative benefits, where the cooperative detector correctly detects a target the receiver misses.

Clear inputs show the same failure stages, and Section~\ref{sec:candidate_diagnosis} reports the fraction of misses with available evidence in each weather group.
This census describes candidate availability on natural inputs. The paired deletion experiment separately isolates the effect of changing the surroundings while holding object measurements fixed.

\begin{table*}[!tbp]
\centering
\caption{Candidate evidence in missed CoInfra detections with V2X-ViT. The three failure stages partition the misses. A identifies misses with an accurate cooperative candidate and B identifies those with a correct receiver detection. These groups can overlap. Cooperative benefits are correctly detected observations that the receiver misses. Counts are target observations across receiving views. Adverse pools rain, snow, and freezing rain.}
\label{tab:supp_natural_evidence}
\small
\setlength{\tabcolsep}{5pt}
\begin{tabular}{@{}lrrrrr@{}}
\toprule
Target observations & Clear & Rain & Snow & Freezing rain & All adverse \\
\midrule
Annotated target observations & 264 & 110 & 226 & 326 & 662 \\
Originally correct & 96 & 59 & 69 & 102 & 230 \\
Missed & 168 & 51 & 157 & 224 & 432 \\
\midrule
Accurate candidates below score threshold & 40 & 22 & 30 & 56 & 108 \\
Accurate candidates suppressed by NMS & 2 & 1 & 5 & 3 & 9 \\
No accurate valid cooperative candidate & 126 & 28 & 122 & 165 & 315 \\
\midrule
Accurate cooperative candidate (A) & 42 & 23 & 35 & 59 & 117 \\
Correct receiver detection (B) & 6 & 3 & 8 & 11 & 22 \\
Overlap of A and B & 4 & 3 & 6 & 8 & 17 \\
Either A or B & 44 & 23 & 37 & 62 & 122 \\
\midrule
Cooperative benefits & 59 & 34 & 48 & 58 & 140 \\
\bottomrule
\end{tabular}
\end{table*}

\subsection{Quality Selection with Fixed Original Candidates}
\label{supp:fixed_selection}
The candidate census identifies missed targets for which accurate boxes remain available.
To test whether current point measurements help use those boxes, we keep all original candidates fixed and train only $R_0$, corresponding to $\lambda=0$.
We compare native scores with quality regressors using box attributes alone or augmented with point statistics from the receiver or both sources.
The learned variants share the same 402,139 original-candidate training examples, overlap targets, date weights, and 300-round regressor settings from Section~\ref{supp:quality_training}.
Unused point-statistic entries are zero during both fitting and inference.
Seeds 20261019, 20261027, and 20261028 change only the regressor's random state.
No correction head is fitted for this comparison, and all variants use identical candidate geometry, budgets, and post-processing.

Table~\ref{tab:fixed_selection} reports adverse-weather AP70, retention, and false detections. Table~\ref{tab:supp_fixed_selection}(a) gives adverse-weather AP50 and clear-weather AP70.
Both-source AP70 is 29.37, 29.22, and 29.23 across the three fits, compared with 26.47, 26.56, and 26.65 for box attributes.
Receiver statistics provide an intermediate improvement, and adding peer statistics increases adverse AP70 in every fit.
Both-source statistics also improve clear-input AP.
Of the 108 adverse-weather misses with accurate candidates below the score threshold, both-source statistics recover seven, seven, and eight.
They recover two, one, and one of the nine misses with above-threshold accurate candidates that were not retained.
These recoveries show that the quality estimates can bring some available accurate boxes back into the final detections.

On the controlled pairs, receiver statistics recover five induced misses per fit but lose two previously stable observations.
Both-source statistics recover four to five while retaining all 119 stable observations (Table~\ref{tab:supp_fixed_selection}(b)).
Together, these results show that both-source statistics help preserve stable detections as well as recover induced misses, beyond the box-attribute regressor used here.

\begin{table*}[!tbp]
\centering
\caption{Quality Selection with fixed original V2X-ViT candidates. Learned variants use $\lambda=0$ and share the training candidate bank, targets, regressor capacity, and evaluation rules. AP values are multiplied by 100 and reported as three-fit mean with sample SD. Counts are fit means. The native-score reference is one fixed output. In (b), recovery requires correctness on both the original and exterior-deleted inputs, following 31 induced losses and 119 stable observations.}
\label{tab:supp_fixed_selection}
\small
\setlength{\tabcolsep}{5pt}
(a) Natural inputs\par\smallskip
\begin{tabular}{@{}lrr@{}}
\toprule
Scoring evidence & Adverse AP50 & Clear AP70 \\
\midrule
Original scores & 46.60 & 26.85 \\
Box attributes & 47.81$\pm$0.06 & 26.39$\pm$0.24 \\
+ Receiver points & 51.45$\pm$0.03 & 29.26$\pm$0.27 \\
+ Both sources & \textbf{53.91}$\pm$0.10 & \textbf{29.81}$\pm$0.23 \\
\bottomrule
\end{tabular}
\par\medskip
(b) Controlled exterior removal\par\smallskip
\begin{tabular}{@{}lrrrr@{}}
\toprule
Scoring evidence & AP70 & Rescued & New losses & Errors \\
\midrule
Original scores & 40.84 & 0.00 & \textbf{0.00} & 31.00 \\
Box attributes & 41.07$\pm$0.05 & 1.00 & 9.33 & 39.33 \\
+ Receiver points & 42.56$\pm$0.06 & \textbf{5.00} & 2.00 & 28.00 \\
+ Both sources & \textbf{44.38}$\pm$0.03 & 4.67 & \textbf{0.00} & \textbf{26.33} \\
\bottomrule
\end{tabular}
\end{table*}

\subsection{Candidate Repair on Natural Inputs}
\label{supp:default_attribution}

Having isolated selection with fixed original boxes, we next examine what AWCR candidate repair adds on the same natural inputs.
We compare AWCR at $\lambda=0.7$ with original-candidate selection at $\lambda=0$ for each of the three fits, using score $>0.27$ and BEV overlap $\geq0.7$.
Gains are targets detected only by AWCR, and losses are targets detected only with $\lambda=0$.
Table~\ref{tab:supp_default_attribution} reports these changes for adverse-weather inputs.

\begin{table}[!tbp]
\centering
\caption{AWCR versus original-candidate selection for the same fixed V2X-ViT detector on 662 natural adverse-weather target observations. Gains requiring new geometry satisfy both the full-pool and fixed-score checks. False detections use the same score threshold as correct detections. Fits follow Table~\ref{tab:supp_controlled_bridge}.}
\label{tab:supp_default_attribution}
\small
\setlength{\tabcolsep}{4pt}
\begin{tabular}{@{}lrrr@{}}
\toprule
Outcome & Fit 1 & Fit 2 & Fit 3 \\
\midrule
Gained correct observations & 14 & 15 & 13 \\
Lost correct observations & 6 & 6 & 8 \\
Gains requiring new geometry & 4 & 2 & 2 \\
False detections, $\lambda=0$ & 264 & 260 & 261 \\
False detections, AWCR & 288 & 289 & 294 \\
\bottomrule
\end{tabular}
\end{table}

To identify gains that require new geometry, we first search all valid car boxes produced by the original detector, including those outside the admission budget.
We then replace each admitted corrected box with the original box from the same anchor, keeping candidate identities, branch labels, and final confidence scores unchanged.
Using the same current points and maps, we recompute foreground support for the replacement boxes, then rerun non-maximum suppression and matching.
We count a gain as requiring new geometry if the selected box comes from the corrected branch, no valid original candidate is accurate, and the target becomes a miss after this replacement.

The adverse-weather gains that meet these checks span five distinct target observations at five physical times.
The best original overlaps range from 0.6852 to 0.699993, and the selected corrected boxes reach 0.7002 to 0.7433.
Reverting the corrected coordinates leaves each of these targets without an accurate admitted candidate, showing that the local geometry changes carry the boxes across the evaluation threshold.

On natural clear inputs, AWCR gains 8, 10, and 6 correct observations and loses 2, 2, and 3 across fits.
New geometry accounts for 5, 5, and 3 of those gains.

To understand the corresponding losses, we trace the accurate candidates admitted by AWCR.
Across both weather groups, all 27 losses counted over the three fits retain an accurate original candidate there.
NMS suppresses an above-threshold accurate candidate in 25 cases. In the other two, the pool has no eligible accurate candidate above the score threshold.
On adverse inputs, AWCR improves AP70 over selection at $\lambda=0$ while producing more false detections and losing some detections kept by selection at $\lambda=0$.

\subsection{Candidate Repair across CoInfra Recordings}
\label{supp:coinfra_recordings}

We evaluate the same frozen V2X-ViT detector and three AWCR fits at $\lambda=0.7$ on each adverse-weather recording (Table~\ref{tab:supp_coinfra_recordings}). The four development recordings use the evaluation times in Table~\ref{tab:supp_real_split}. For a test on a separate date, we use the entire November 27 heavy-snow slice6. The detector, repair components, correction weight, and background maps were fixed before this evaluation. November 27 was absent from their training, selection, and calibration data. Earlier project work used slices 0--5 from that date.

All 581 consecutive timestamps in slice6 meet the existing 100\,ms four-node synchronization criterion (maximum observed offset 6\,ms). Each time contributes both receiving directions for pairs $(6,9)$ and $(7,8)$. The 2,324 views include four with no visible car. The 7,506 visible car observations span 15 annotation IDs over 58 seconds.

\begin{table*}[!tbp]
\centering
\caption{Paired car BEV AP70 for the same V2X-ViT detector with and without AWCR on natural CoInfra inputs. Car observations count each visible annotation once per time and receiving view. The With AWCR column gives the mean and sample SD across the three fixed fits. Each $\Delta$ is measured against the original detector on the same views. Pooled rows rank all predictions jointly before computing each fit's AP.}
\label{tab:supp_coinfra_recordings}
\small
\setlength{\tabcolsep}{4pt}
\begin{tabular}{@{}lrrrrrrrrr@{}}
\toprule
Recording & Times & Views & Car obs. & Baseline & With AWCR & $\Delta$ Fit 1 & $\Delta$ Fit 2 & $\Delta$ Fit 3 & Mean $\Delta$\\
\midrule
Feb.\ 12, snow & 12 & 48 & 134 & 23.79 & 30.10$\pm$0.24 & +6.22 & +6.58 & +6.13 & +6.31\\
Mar.\ 24, rain & 12 & 48 & 110 & 45.78 & 50.50$\pm$0.34 & +4.35 & +5.01 & +4.80 & +4.72\\
Apr.\ 2, freezing rain & 32 & 128 & 326 & 22.51 & 27.05$\pm$0.48 & +4.35 & +5.08 & +4.17 & +4.53\\
Apr.\ 7, snow & 8 & 32 & 92 & 20.21 & 20.62$\pm$0.86 & +1.28 & +0.38 & $-0.43$ & +0.41\\
Nov.\ 27, snow (slice6) & 581 & 2,324 & 7,506 & 23.42 & 27.13$\pm$0.17 & +3.58 & +3.90 & +3.64 & +3.71\\
\midrule
Four development recordings & 64 & 256 & 662 & 25.70 & 29.92$\pm$0.34 & +4.25 & +4.54 & +3.86 & +4.22\\
All five recordings & 645 & 2,580 & 8,168 & 23.55 & 27.34$\pm$0.18 & +3.68 & +4.00 & +3.70 & +3.79\\
\bottomrule
\end{tabular}
\end{table*}

The mean gain is positive on each recording. The smallest gain is on April 7, where the third fit loses 0.43 points. Omitting each development recording in turn and rescoring the other three with the same fits gives mean gains of 3.69, 4.08, 3.80, and 5.10 points in table order. Slice6 supplies 90.1\% of the views in the five-recording pool. The reported SD measures variation across repair fits on fixed inputs. Temporal repetition and the small number of recordings limit conclusions about variation across weather events.

\section{Repair Components and Operating Conditions}
\label{supp:components}

\begin{table*}[!tbp]
\centering
\caption{V2X-ViT component and training comparisons. Natural adverse-weather columns use 662 target observations, with Kept following 230 originally correct detections. Clear AP70 uses 264 observations. Controlled repair pairs contain 31 induced misses and 119 stable observations in the 64-view clear-input cohort. Natural comparisons use shared support filtering, while controlled comparisons use native post-processing. (a) Matched ablations at $\lambda=0.7$ report three-fit mean AP with sample SD and mean counts. (b) Structural and training variants at $\lambda=1$ each use a single fit. Rescued requires correctness on both paired inputs, and New losses counts originally stable observations missed on either. Bold marks column optima within each panel.}
\label{tab:supp_full_components}
\small
\setlength{\tabcolsep}{3pt}
(a) Matched component ablations ($\lambda=0.7$)\par\smallskip
\begin{tabular}{@{}lrrrrrrr@{}}
\toprule
& \multicolumn{3}{c}{Natural adverse weather} & Clear & \multicolumn{3}{c}{Controlled repair pairs} \\
\cmidrule(lr){2-4}\cmidrule(lr){5-5}\cmidrule(l){6-8}
Configuration & AP70 & Kept & False & AP70 & AP70 & Rescued & New losses \\
\midrule
Original & 25.70 & \textbf{230.00} & 447.00 & 26.85 & 40.84 & 0.00 & \textbf{0.00} \\
AWCR w/o Point Statistics & 28.24$\pm$0.11 & 224.00 & 392.33 & 28.61$\pm$0.66 & 42.84$\pm$0.40 & 7.33 & 4.33 \\
AWCR w/o $\mathcal L_{\mathrm{keep}}$ & 29.83$\pm$0.35 & 225.67 & \textbf{286.67} & 30.60$\pm$0.25 & 44.74$\pm$0.35 & \textbf{8.00} & 2.33 \\
AWCR & \textbf{29.92}$\pm$0.34 & 225.33 & 290.33 & \textbf{30.64}$\pm$0.49 & \textbf{44.88}$\pm$0.20 & 7.33 & 1.33 \\
\bottomrule
\end{tabular}
\par\medskip
(b) Structural and training comparisons, single fits at $\lambda=1$\par\smallskip
\begin{tabular}{@{}lrrrr@{}}
\toprule
& \multicolumn{2}{c}{Natural adverse weather} & \multicolumn{2}{c}{Controlled repair pairs} \\
\cmidrule(lr){2-3}\cmidrule(l){4-5}
Variant & AP70 & False & Rescued & New losses \\
\midrule
Corrected candidates only & 28.69 & 360 & \textbf{12} & 10 \\
Without Quality Selection & 28.71 & 566 & 11 & 8 \\
\midrule
Training with original patches & \textbf{30.72} & 336 & 11 & 10 \\
Training with random exterior loss & 29.01 & \textbf{334} & 9 & 7 \\
Training with directional exterior loss & 30.37 & 335 & 11 & \textbf{6} \\
\bottomrule
\end{tabular}
\end{table*}

\subsection{Correction Weight across Models and Recordings}
\label{supp:correction_trust}
With the repair models fixed, the CoInfra-selected $\lambda=0.7$ rescues fewer V2X-ViT misses and introduces fewer new losses than $\lambda=1$ on the controlled pairs, while $\lambda=0$ gives the highest AP70 on the dry DAIR/V2X-Seq test recording (Tables~\ref{tab:supp_awcr_repeated} and~\ref{tab:supp_natural_rain}).
In a single AttFuse repair fit on natural adverse-weather CoInfra inputs, the default weight raises car AP70 from 10.55 to 15.84 and false detections from 57 to 143, compared with 16.11 AP70 and 215 false detections at $\lambda=1$.

\subsection{Component and Training Comparisons}
Table~\ref{tab:supp_full_components}(a) removes one component at a time at the default weight $\lambda=0.7$. It tests point statistics in Quality Selection and the keep loss, which discourages changes to originally correct predictions during Local Correction training.
To remove point statistics, we zero all 60 point entries in both quality regressors during fitting and inference. This leaves the correction inputs and shared support filter fixed.
To test the keep loss, we set its coefficient to zero while keeping Local Correction's labels, bounds, initialization, order, and update budget fixed. We refit $R_1$ on the resulting candidates and reuse $R_0$.
Removing the keep loss lowers mean adverse AP70 by 0.09 points.

Panel (b) uses one fit per row at $\lambda=1$, with the full model in the last row.
On 64 controlled views, keeping original boxes alongside corrected boxes reduces new losses from 10 to 6 and paired rescues from 12 to 11.
In natural adverse weather, Quality Selection reduces false detections from 566 to 335 and raises AP70 from 28.71 to 30.37.
The point-statistics contribution is isolated by the three-fit ablation in panel (a) and the fixed-original-candidate comparison in Table~\ref{tab:supp_fixed_selection}.

The three training rows in panel (b) change the inputs assigned to the constructed exterior-loss conditions. These conditions use original patches, equal-count random exterior loss (seed 20261025), or measured directional loss. The original observations, training objectives, order, and update budgets stay fixed.
Training without constructed loss gives the highest adverse AP70 in this comparison, while directional loss gives the fewest new losses on controlled pairs.
Natural clear AP70 is 31.26, 31.66, and 32.65 when those conditions use original patches, random exterior loss, and directional exterior loss, respectively.

A separate single-fit comparison at $\lambda=1$ changes only which source statistics enter Quality Selection, with candidates, box attributes, targets, and fitting settings fixed.
\phantomsection\label{supp:quality_sources}
Both views still feed detection, correction, and natural support filtering.
Using both sources produces 47 fewer false detections and one fewer correct detection than using receiver statistics alone in natural adverse weather, while reducing controlled errors from 28 to 26.
Receiver statistics alone give the highest natural clear AP70 in this comparison.

\section{Training References and Weather Transfer}
\label{supp:training_references}\label{supp:additional_results}

\begin{table*}[!tbp]
\centering
\begin{minipage}{\textwidth}
\centering
\captionof{table}{Full-detector references and candidate repair. V2X-ViT + AWCR uses a single fit (seed 20261019). Table~\ref{tab:supp_awcr_repeated} reports AWCR on Robo3D-DIT-adapted V2X-ViT. (a) Natural-input columns use shared support filtering, while controlled columns use native post-processing on the 64-view clear-input repair cohort after exterior deletion. (b) Correct, False, and Retained refer to natural adverse weather, with retention following 230 native correct observations. Controlled recovery follows 31 induced losses and 119 stable observations. Errors sums unrecovered induced misses and new losses across the paired inputs. Bold marks column optima. AP values are multiplied by 100.}
\label{tab:supp_external_references}\label{tab:supp_additional_metrics}
\small
\setlength{\tabcolsep}{4pt}
(a) Detection across evaluation conditions\par\smallskip
\begin{tabular}{@{}lrrrrr@{}}
\toprule
Method & Adverse mAP50 & Adverse mAP70 & Adverse car AP70 & Clear car AP70 & Controlled car AP70 \\
\midrule
Original & 12.08 & 6.48 & 25.70 & 26.85 & 40.84 \\
DSOR (default) & 7.28 & 2.51 & 9.76 & 14.10 & 25.41 \\
Ordinary fine-tuning & 15.73 & 6.46 & 25.16 & 41.31 & 42.54 \\
RCP-Mix (LiDAR) & 16.90 & 7.17 & 27.55 & \textbf{44.31} & 43.88 \\
Robo3D-DIT (adapt.) & \textbf{17.02} & \textbf{7.71} & 29.11 & 40.12 & \textbf{46.50} \\
V2X-ViT + AWCR & 14.34 & 7.54 & \textbf{29.95} & 30.90 & 45.05 \\
\bottomrule
\end{tabular}

\par\smallskip
(b) Natural retention and controlled recovery\par\smallskip
\begin{tabular}{@{}lrrrrrr@{}}
\toprule
& \multicolumn{3}{c}{Natural adverse weather} & \multicolumn{3}{c}{Controlled pairs} \\
\cmidrule(lr){2-4}\cmidrule(l){5-7}
Method & Correct & False & Retained & Rescued & New & Errors \\
\midrule
Original & 230 & 447 & \textbf{230} & 0 & \textbf{0} & 31 \\
DSOR (default) & 98 & \textbf{257} & 80 & 3 & 44 & 72 \\
Ordinary fine-tuning & 216 & 295 & 152 & 9 & 21 & 43 \\
RCP-Mix (LiDAR) & 245 & 308 & 173 & \textbf{15} & 21 & 37 \\
Robo3D-DIT (adapt.) & \textbf{264} & 258 & 186 & \textbf{15} & 22 & 38 \\
V2X-ViT + AWCR & 246 & 288 & 227 & 8 & 1 & \textbf{24} \\
\bottomrule
\end{tabular}
\end{minipage}
\par\smallskip

\end{table*}

\subsection{Full-Detector Training References}

We compare candidate repair with retraining the full detector from the same CoInfra-adapted V2X-ViT checkpoint.
Table~\ref{tab:supp_external_references} extends Table~\ref{tab:detection} to natural clear inputs, controlled exterior removal, and retention of originally correct detections.
AWCR trains only its repair components on patches. By contrast, ordinary fine-tuning and RCP-Mix update the full detector with native five-class supervision for 3,000 steps. Both use four observations per step in a fixed order and follow the optimizer settings in Section~\ref{supp:local_training}.

The LiDAR adaptation of RCP-Mix applies Eq.~(6) of the original method at the scatter and shrink outputs with $\operatorname{Beta}(0.1,0.1)$ weights and variance epsilon $10^{-5}$.
At inference, this LiDAR adaptation uses the detector's stored normalization statistics, whereas the published camera method also uses online adaptation.

The Robo3D-DIT~\cite{kong2023robo3d} adaptation instead trains two branches initialized from the same checkpoint. At each step, both receive the same four original observations, but the second branch uniformly retains only $\lfloor0.6N\rfloor$ of each source cloud's $N$ points without replacement.
The joint objective supervises detection in both branches with the native five-class labels and aligns their features.
\begin{equation}
\begin{aligned}
\mathcal L_{\mathrm{DIT}}={}&\mathcal L_{\mathrm{det}}(F)+\mathcal L_{\mathrm{det}}(P)\\
&+50\,\underset{x\in\mathcal O_F,\,c}{\operatorname{mean}}(F_c(x)-P_c(x))^2\\
&+100\,\underset{x\in\mathcal O_P,\,c}{\operatorname{mean}}(F_c(x)-P_c(x))^2.
\end{aligned}
\label{eq:supp_dit_training}
\end{equation}

Here $F,P\in\mathbb R^{256\times144\times144}$ are the pre-head fused features of the full and thinned branches, respectively, and $c$ indexes channels. For each branch, its occupied-cell set is the union of cells occupied by either source, with $\mathcal O_P\subseteq\mathcal O_F$.
Both branches train for 3,000 steps with separate normalization statistics on the native grid, with flexible voxelization disabled. The two-branch training adds computation on full inputs, while inference uses only the final weights of the thinned branch.

These training choices produce different tradeoffs. Among the fixed runs, RCP-Mix gives the highest AP70 on natural clear inputs, while Robo3D-DIT gives the highest adverse-weather four-class mAP and controlled Car AP70.
AWCR instead retains more native correct detections in adverse weather and loses fewer previously stable targets on the controlled pairs than either method.

All detection references use the common post-processing from Section~\ref{supp:evaluation}. For the original detector, its shared support filter raises adverse-weather Car AP70 from 24.46 to 25.70 and four-class mAP70 from 6.17 to 6.48.
The point-filtering references act before detection. DROR uses angular resolution $0.16^\circ$, radius multiplier three, minimum radius 0.04\,m, and three neighbors including the query.
DSOR~\cite{kurup2021dsor} uses the author's defaults of five neighbors, standard-deviation multiplier 0.01, and range multiplier 0.05.
DROR and DSOR process each sensor independently and preserve the values and order of retained points. The support mask comes from the original current input in both cases.
Late fusion runs detection separately at the two sensors and merges their boxes using the original confidence scores, without score calibration. It shares the total candidate budget and suppression rule with the cooperative detector.

\begin{table*}[!tbp]
\centering
\begin{minipage}{\textwidth}
\centering
\captionof{table}{AWCR configurations on the original and Robo3D-DIT-adapted V2X-ViT detectors. (a) Natural inputs use shared support filtering, while controlled inputs use native post-processing after exterior deletion in the 64-view clear-input repair cohort. Repair AP is mean$\pm$sample SD over three fits. Only Car predictions change. For the original detector, dividing the unrounded mean adverse-weather Car AP70 gain by four gives a 1.05-point gain in four-class mAP70 after rounding. (b) Counts are fit means for repair variants. Natural adverse-weather retention follows 230 native correct observations for the original detector and 264 for the adapted detector. On the shared controlled cohort, these detectors have 31 and 19 induced losses and 119 and 122 stable observations, respectively. Errors sums unrecovered induced misses and new losses across the paired inputs. Each repair fit uses its own quality regressors, shared across its weight configurations. AWCR uses the default $\lambda=0.7$.}
\label{tab:supp_awcr_repeated}\label{tab:supp_seeds}\label{tab:supp_robust_repair}
\small
\setlength{\tabcolsep}{5pt}
(a) Detection\par\smallskip
\begin{tabular}{@{}lrrrr@{}}
\toprule
Method & Adverse AP50 & Adverse AP70 & Clear AP70 & Controlled AP70 \\
\midrule
\multicolumn{5}{l}{Original detector} \\
Original & 46.60 & 25.70 & 26.85 & 40.84 \\
$\lambda=0$ & 53.91$\pm$0.10 & 29.27$\pm$0.08 & 29.81$\pm$0.23 & 44.38$\pm$0.03 \\
$\lambda=1$ & 56.55$\pm$0.28 & 29.94$\pm$0.96 & 32.90$\pm$1.72 & 43.00$\pm$0.91 \\
AWCR & 55.90$\pm$0.28 & 29.92$\pm$0.34 & 30.64$\pm$0.49 & 44.88$\pm$0.20 \\
\multicolumn{5}{l}{Robo3D-DIT-adapted V2X-ViT} \\
Original & 57.76 & 29.11 & 40.12 & 46.50 \\
$\lambda=0$ & 60.68$\pm$0.16 & 30.82$\pm$0.11 & 41.01$\pm$0.02 & 47.99$\pm$0.13 \\
$\lambda=1$ & 62.98$\pm$0.18 & 31.71$\pm$0.32 & 42.62$\pm$0.50 & 48.98$\pm$0.02 \\
AWCR & 61.60$\pm$0.40 & 31.47$\pm$0.14 & 41.42$\pm$0.15 & 47.80$\pm$0.27 \\
\bottomrule
\end{tabular}

\par\smallskip
(b) Retention and recovery\par\smallskip
\begin{tabular}{@{}lrrrrrr@{}}
\toprule
& \multicolumn{3}{c}{Natural adverse weather} & \multicolumn{3}{c}{Controlled pairs} \\
\cmidrule(lr){2-4}\cmidrule(l){5-7}
Method & Correct & False & Retained & Rescued & New losses & Errors \\
\midrule
\multicolumn{7}{l}{Original detector} \\
Original & 230.00 & 447.00 & 230.00 & 0.00 & 0.00 & 31.00 \\
$\lambda=0$ & 237.33 & 261.67 & 228.67 & 4.67 & 0.00 & 26.33 \\
$\lambda=1$ & 247.00 & 348.33 & 215.33 & 9.33 & 5.33 & 27.00 \\
AWCR & 244.67 & 290.33 & 225.33 & 7.33 & 1.33 & 25.00 \\
\multicolumn{7}{l}{Robo3D-DIT-adapted V2X-ViT} \\
Original & 264.00 & 258.00 & 264.00 & 0.00 & 0.00 & 19.00 \\
$\lambda=0$ & 269.67 & 259.67 & 263.33 & 2.00 & 0.00 & 17.00 \\
$\lambda=1$ & 282.33 & 334.00 & 249.33 & 4.67 & 1.00 & 15.33 \\
AWCR & 280.67 & 290.67 & 259.33 & 3.00 & 0.00 & 16.00 \\
\bottomrule
\end{tabular}
\end{minipage}
\par\smallskip

\end{table*}

\subsection{Repeated Repair Fits}
\label{supp:seed_stability}\label{supp:robust_default}

For each detector (original or Robo3D-DIT-adapted V2X-ViT), we fit three repair heads, each with its own pair of quality regressors. We evaluate all six fits on the same observations (Table~\ref{tab:supp_awcr_repeated}).
Seeds 20261019, 20261027, and 20261028 change head initialization and the order of the 3,000 minibatches while preserving the four patches in each minibatch. The detector, data split, patches, normalization, supervision, and total updates stay fixed.

\begin{samepage}
Each repair fit has a 300-round $R_0$ regressor trained on original candidates and a 300-round $R_1$ regressor trained on both branches. For each fixed detector, the three $R_0$ fits share its fixed original-candidate training bank. For each detector and seed, the fitted weights stay fixed across both endpoints and the default configuration.
\par
\end{samepage}

Table~\ref{tab:supp_awcr_repeated} compares both endpoints and the default configuration across all fits.

The Robo3D-DIT-adapted detector retains the V2X-ViT architecture but starts from stronger predictions. At $\lambda=0.7$, we evaluate the three repair fits trained on this adapted detector. Candidate admission, weighted scoring, support filtering, and suppression are recomputed at this weight. AWCR raises adverse-weather AP70 from 29.11 to $31.47\pm0.14$ and exceeds both the fixed detector and selection at $\lambda=0$ in all three fits.

The additional detections come with a retention tradeoff. Compared with selection at $\lambda=0$, the mean correct count increases by 11 and the false count by 31, while four fewer native correct observations are retained. Relative to the fixed detector's predictions, AWCR gains 23, 20, and 21 correct observations and loses 7, 2, and 5 in seed order. Compared with $\lambda=1$, it retains ten more native correct observations and produces 43.33 fewer false detections on average, with slightly lower AP70.

On the adapted detector, AWCR raises mean snow AP70 from 26.18 to 27.91 over selection alone ($\lambda=0$). The mean difference is $+0.03$ points in rain (negative in two fits) and $-0.03$ in freezing rain.
On the controlled pairs, AWCR recovers three of the 19 induced losses in each fit, compared with two for $\lambda=0$. Both preserve all 122 originally stable observations, reducing the error count from 17 to 16. Yet controlled AP70 decreases from 47.99 to 47.80, because the ranking of all predictions and recovery of this fixed target cohort measure different aspects of repair.

\subsection{Training Repair without the Tested Weather}
\label{supp:weather_transfer}

\begin{table*}[!tbp]
\centering
\begin{minipage}{\textwidth}
\centering
\captionof{table}{Leave-one-weather-out repair with a fixed V2X-ViT detector. (a) Car AP70. Validation inputs from the remaining weather conditions select $\lambda$ of $0.8$, $0.4$, and $0.7$ for rain, snow, and freezing rain, respectively. Each fold uses its selected weight across fits. (b) Pooled AP50, counts at score $>0.27$, and correct detections at equal false counts. The counts use overlap $\geq0.7$ for a match. Pooled AP ranks predictions jointly across the three weather test folds. For repair variants, AP is shown as mean$\pm$sample SD and counts as means over three fits. The inherited weight of $0.7$ was selected using all training weather.}
\label{tab:supp_weather_transfer}
\small
\setlength{\tabcolsep}{3.5pt}
(a) Detection on held-out repair weather\par\smallskip
\begin{tabular}{@{}lrrrr@{}}
\toprule
Method & Rain & Snow & Freezing rain & Pooled folds \\
\midrule
Original & 45.78 & 21.96 & 22.51 & 25.70 \\
$\lambda=0$ & 50.09$\pm$0.40 & 24.89$\pm$0.10 & 26.03$\pm$0.12 & 29.16$\pm$0.08 \\
$\lambda=1$ & 46.67$\pm$2.69 & 19.62$\pm$1.60 & 28.55$\pm$1.27 & 27.67$\pm$1.05 \\
AWCR (source val.) & 48.25$\pm$1.60 & 25.31$\pm$0.28 & 26.81$\pm$0.68 & 29.71$\pm$0.42 \\
\midrule
AWCR ($\lambda=0.7$, inherited) & 48.12$\pm$1.33 & 26.21$\pm$0.21 & 26.81$\pm$0.68 & 29.74$\pm$0.33 \\
\bottomrule
\end{tabular}

\par\smallskip
(b) Pooled retention and operating-range comparison\par\smallskip
\begin{tabular}{@{}lrrrrrrrr@{}}
\toprule
Method & AP50 & Correct & False & Retained & 64 false & 128 false & 256 false & 512 false \\
\midrule
Original & 46.60 & 230.00 & 447.00 & 230.00 & 137.00 & 174.00 & 219.00 & 236.00 \\
$\lambda=0$ & 53.18$\pm$0.22 & 238.33 & 282.00 & 227.33 & 165.33 & 208.00 & 234.67 & 245.33 \\
$\lambda=1$ & 55.16$\pm$0.20 & 240.67 & 489.67 & 203.00 & 159.00 & 194.00 & 228.67 & 240.67 \\
AWCR (source val.) & 55.07$\pm$0.29 & 239.33 & 307.00 & 222.00 & 178.00 & 214.00 & 234.00 & 248.33 \\
\midrule
AWCR ($\lambda=0.7$, inherited) & 55.70$\pm$0.47 & 247.00 & 365.00 & 219.67 & 179.67 & 209.33 & 237.00 & 250.00 \\
\bottomrule
\end{tabular}
\end{minipage}
\par\smallskip

\end{table*}

To test transfer to a weather condition excluded from repair fitting, we train the repair components and choose $\lambda$ using only the remaining conditions.
We refer to these remaining conditions as the source conditions. The original CoInfra-adapted V2X-ViT detector stays fixed across all tests (Section~\ref{supp:data}).
For the rain, snow, and freezing-rain test folds, the source training sets contain 320, 256, and 320 views with 4,224, 3,840, and 4,224 patches, respectively. The corresponding test sets contain 48, 80, and 128 views with 110, 226, and 326 Car annotations.

We estimate both per-channel feature RMS scales and $180\times90$ angular profiles from source inputs. The profiles use both clear training recordings as the reference and require a mean clear count of at least 1 and an adverse-to-clear ratio below $0.5$.
An identifier hash assigns the source profiles to inputs for exterior deletion, which preserves object interiors. Repair fits then follow the update budget and final-weight rule in Sections~\ref{supp:local_training} and~\ref{supp:seed_stability}.

To choose the correction weight, we divide the source data into 50, 40, and 50 fitting times, 10, 8, and 10 buffer times, and 20, 16, and 20 validation times for the rain, snow, and freezing-rain tests, respectively. Each time contributes four views.
On each fitting subset, we train one head (seed 20261019, 3,000 four-patch updates) and two 300-round quality regressors.
We select a weight from $\{0,0.1,\ldots,1\}$ using AP70 on unmodified validation inputs from those source conditions, favoring lower weights within 0.1 AP points of the maximum.
We fix this weight before testing and use it for all three repair training runs on the full source training set.
Table~\ref{tab:supp_weather_transfer} labels this configuration AWCR (source val.) and reports the inherited weight separately, since that value was chosen using all training weather.

For each seed, we pool predictions across the three weather tests before computing AP70.
With weights chosen on source validation inputs, AWCR exceeds selection at $\lambda=0$ in pooled AP70 in all three runs (Table~\ref{tab:supp_weather_transfer}).
Its mean AP70 is higher in snow and freezing rain, while selection at $\lambda=0$ leads in rain.

To compare the methods at the same false-detection count, we use their complete confidence curves and keep tied-score groups together.
With source-selected weights, AWCR finds more targets than selection at $\lambda=0$ at 128 and 512 false detections and slightly fewer at 256.
The fixed-score counts show the corresponding retention tradeoff, with fewer originally correct detections retained than at $\lambda=0$ and more than at $\lambda=1$.

\section{Public Pretrained Detectors}
\label{supp:public_detectors}

To test whether candidate repair benefits released detectors, we train its components on each dataset's training inputs while keeping the released detector fixed.
The three correction fits use seeds 20261019, 20261020, and 20261021, each with 3,000 updates and four patches per update. Each quality regressor uses 300 boosting rounds.
We evaluate the final weights with the CoInfra-selected default $\lambda=0.7$.
For each detector, Original and + AWCR use the same evaluation inputs and annotations and retain the native decoding and filtering rules.
To compute Vehicle BEV AP, we pool predictions across evaluation inputs and rank them by confidence, except in the frame-order comparison below.
This global ordering follows HEAL's released evaluator and is enabled by default in FadeLead's released inference entry.
We also retain their framewise confidence-ordered matching, VOC precision-envelope integration, annotation rules, and treatment of empty predictions.

\begin{table}[!t]
\centering
\caption{Vehicle BEV AP multiplied by 100. Unmarked rows are published results. We use the released evaluators' global confidence ranking for the reproduced detectors and report mean$\pm$sample SD over three fits for each added head. AWCR uses $\lambda=0.7$ from CoInfra. Both RoI rows use the same refiner with different scoring rules. Table~\ref{tab:detection} reports paired AWCR gains.}
\label{tab:supp_public_benchmarks}\label{tab:supp_public_repeats}
\small
\setlength{\tabcolsep}{6pt}
\begin{tabular}{@{}lrr@{}}
\toprule
Method & AP50 & AP70 \\
\midrule
\multicolumn{3}{@{}l}{V2X-Radar-C, LiDAR-only} \\
Late fusion & 65.75 & 39.37 \\
F-Cooper & 73.44 & 50.04 \\
CoAlign & 80.42 & 60.18 \\
HEAL & 83.00 & 67.57 \\
HEAL (reproduced) & 82.99 & 67.53 \\
\quad + AWCR & 84.21$\pm$0.08 & 68.22$\pm$0.04 \\
\quad + RoI (direct quality) & 84.76$\pm$0.58 & 69.96$\pm$0.77 \\
\quad + RoI (matched mapping) & 83.57$\pm$0.05 & 68.96$\pm$0.12 \\
\midrule
\multicolumn{3}{@{}l}{DAIR-V2X-C} \\
FadeLead hybrid & 79.66 & 66.51 \\
FadeLead hybrid (reproduced) & 79.68 & 66.53 \\
\quad + AWCR & 80.31$\pm$0.07 & 67.24$\pm$0.04 \\
\bottomrule
\end{tabular}
\end{table}

\subsection{HEAL on V2X-Radar-C}
We evaluate HEAL~\cite{lu2024heal} on all 922 validation times across five recordings with 19,704 Vehicle annotations. Table~\ref{tab:supp_public_benchmarks} lists the LiDAR-only comparisons from the V2X-Radar model zoo~\cite{yang2025v2xradar}\footnote{\url{https://github.com/yanglei18/V2X-Radar\#models-zoo}}.

Each of the four training recordings contributes 24 fitting times and 12 temporally separated times for weight validation. Our generic region-of-interest (RoI) refiner uses the same 576 training patches, update schedule, and 4,096-candidate budget as AWCR. It predicts box offsets and quality from a rotated $7\times7$ grid of fused BEV features and up to 512 source-tagged points per candidate. Its neural head has 3,460,104 parameters. AWCR uses a 64,209-parameter correction head and two tree regressors. We preset both scoring rules for the refined boxes, using predicted quality alone for the primary comparison and AWCR's score mapping for the second.

Selecting the weight on the 48 held-out times from HEAL's training recordings gives $\lambda=0$ and yields 83.33 AP50 and 67.90 AP70 on the full synchronous validation set. At this weight, the three correction fits share the original-candidate regressor $R_0$ and its predictions.
Transferring the fixed CoInfra weight ($\lambda=0.7$) raises vehicle AP70 from 67.53 to 68.22, improving on the reproduced HEAL baseline in every fit and exceeding the published HEAL result of 67.57 by 0.65 points (Table~\ref{tab:supp_public_repeats}).

\subsection{FadeLead on DAIR-V2X-C}
We evaluate FadeLead~\cite{wu2026fadelead} on all 1,789 validation inputs with the authors' complemented annotations. The primary comparison uses their highest-accuracy 10\% hybrid, which combines intermediate fusion and output merging. We also test the 1\% configuration to examine repair under a lower native foreground-selection ratio. Each retains its annotation filtering, which uses all three XYZ coordinates for the hybrid and only horizontal XY coordinates for the 1\% configuration.

For each configuration, we use 96 training times disjoint from validation. The hybrid provides 192 receiving views and 1,152 patches. The 1\% setting provides 96 views and 576 patches. The two receiving directions of the hybrid share a 64,209-parameter correction head and both quality regressors. We fit in FP32 using the detector's cooperative features and outputs.

The reproduced hybrid closely matches the published baseline, providing the reference for the paired repair comparison in Table~\ref{tab:supp_public_benchmarks}. In the 1\% setting, AWCR raises AP50 from 79.10 to $79.35\pm0.03$ and AP70 from 64.42 to $65.03\pm0.03$ across the same three seeds.

These 10\% and 1\% labels describe FadeLead's native foreground selection. Candidate repair adds communication by transmitting a $64\times100\times252$ FP32 feature map of 6,451,200 bytes per direction and XYZ coordinates at 12 bytes per point.
In the hybrid, these representations add an average of 13,874,506 bytes per input across both directions. Extra boxes and scores use 32 bytes per candidate and add an average of 467 to 508 bytes per input across fits. The total is about 13.875\,MB per input. The 1\% setting adds about 6.864\,MB of representations in one direction. All payload sizes are uncompressed.

\subsection{Candidate Repair with V2X-DGW}
\label{supp:native_dgw}
On OPV2V-W, we evaluate AWCR with the released V2X-DGW checkpoint~\cite{li2025v2xdgw} on all 2,170 receiving inputs per weather. The three repair fits use 192 clean times from 32 scenes disjoint from testing and follow the loss and bounds in Section~\ref{supp:local_training}.

We retain up to seven native agents within 70\,m. To pool the peer inputs, we weight features by occupancy, sum point counts, and merge current points. The 72,401-parameter correction head uses 384 context channels on a $100\times352$ grid.
At inference, we use a candidate budget of 4,096, a score floor of 0.20, and an NMS overlap of 0.15. All configurations use the released annotation range $[-100,100]\times[-40,40]\times[-5,3]$\,m.

Table~\ref{tab:supp_dgw_order} compares AWCR with Original under global confidence ranking and the released evaluator's frame-loading order. Both evaluations use identical predictions, confidence-ordered matching within each frame, and AP interpolation. Only the order used to accumulate matches and false detections differs.
AWCR improves globally ranked AP70 in every fit and weather, but gives lower mean AP70 under frame-order accumulation.

\begin{table}[!tbp]
\centering
\caption{Original and AWCR with the fixed released V2X-DGW checkpoint on all 2,170 OPV2V-W inputs per weather. AP70 is multiplied by 100. Both orders use identical predictions. AWCR uses $\lambda=0.7$ and reports mean$\pm$sample SD across three fits. Changes are computed before rounding.}
\label{tab:supp_native_dgw}\label{tab:supp_dgw_order}
\small
\setlength{\tabcolsep}{4pt}
\begin{tabular}{@{}llrrr@{}}
\toprule
Order & Weather & Original & AWCR & Change \\
\midrule
Global & Fog & 70.06 & 70.41$\pm$0.05 & $+0.35$ \\
& Rain & 75.44 & 75.63$\pm$0.04 & $+0.18$ \\
& Snow & 63.48 & 63.79$\pm$0.12 & $+0.31$ \\
\midrule
Frame & Fog & 65.58 & 62.78$\pm$0.21 & $-2.79$ \\
& Rain & 69.33 & 65.36$\pm$0.15 & $-3.98$ \\
& Snow & 59.56 & 55.91$\pm$0.30 & $-3.65$ \\
\bottomrule
\end{tabular}
\end{table}

The confidence curves help explain the difference between the two evaluations. At 50\% recall, globally ranked AWCR predictions contain fewer false detections than Original in every fit and weather. Once all retained predictions are included, false detections increase in every condition, and correct rain detections fall from 25,773 to between 25,747 and 25,769.
Frame-order accumulation preserves the input sequence, allowing a low-confidence false detection in an early frame to precede a high-confidence match in a later frame. With this ordering, AWCR has lower mean AP70 in all three weather conditions.

\section{Independent Natural-Weather Recordings}
\label{supp:independent_real}\label{supp:natural_rain_recording}

\subsection{Recordings and Detector}

We use independent recordings to examine the mechanism with another detector and test repair beyond its training recording. To pair the data, we match DAIR-V2X-C point clouds by timestamp to V2X-Seq-SPD annotations and calibration~\cite{yu2022dairv2x,yu2023v2xseq}.
Candidate repair trains on 32 uniformly sampled times from dry recording \texttt{0000}. It is evaluated on all 207 eligible pairs from dry recording \texttt{0003} at the same intersection and all 168 from recording \texttt{0042} at another intersection. All three recordings come from the same day.

Lens droplets, water on vehicles, and wet roads identify recording \texttt{0042} as rain-affected. It spans 17\,s and contains 994 observations across 29 tracks. The dry test recording contains 2,820 observations across 59 tracks.

We hold the HEAL DAIR LiDAR F-Cooper checkpoint~\cite{lu2024heal} fixed and retain its native intensity preprocessing, communication range, decoding, and $128\times256$ grid. The vehicle-frame evaluation range is $[-10,79.12]\times[-49.68,49.68]\times[-3,1]$\,m. The Vehicle evaluation category groups Car, Van, Truck, and Bus.

\subsection{Controlled Mechanism Replication}
\label{supp:fcooper_restoration}

For the controlled diagnosis, we transfer the fixed CoInfra freezing-rain angular masks from nodes 7 and 8 to the receiver and peer, respectively. We apply them to the same 375 evaluation inputs without fitting a repair model.
We protect all stored points inside valid annotated 3D boxes, including categories outside the evaluated Vehicle class, and preserve retained coordinates, intensities, and point order. Neighboring exterior removal can nevertheless change voxel counts and derived encodings.
Equal-count random deletion provides a comparison with the same per-source removal counts and three fixed seeds, 20261024 through 20261026. The native communication rule excludes the peer in 30 dry inputs, which remain in the evaluation.

We then test local feature restoration immediately before max fusion, after the native warp into the receiver frame. The intervention uses all available source-representation channels on a $128\times256$ grid covering $[-102.4,102.4]\times[-51.2,51.2]$\,m with 0.8\,m cells, while predictions use the evaluation range above.
The object mask is the union of annotated BEV boxes expanded by half a cell diagonal. We restore features in this mask from the original input and continue the unchanged detector.
For exterior controls, we rotate the object-mask cell centers by $90^\circ$, $180^\circ$, and $270^\circ$ around the receiver origin. On the rectangular grid, we remove their overlap with the object mask and fill each to the same area using the remaining cells closest to the rotated centers. Each mask covers 312.02\,m$^2$ on average. The controls match area, while distance, occupancy, and feature-change magnitude may differ. For comparison, the object region has mean range 51.70\,m and root-mean-square feature change 0.0209, compared with 52.65\,m and 0.0201 for the $180^\circ$ control.
To check the intervention implementation, we replay unchanged features and replace the entire feature tensor. Both reproduce the original dense output on every input, and the original final predictions reproduce the earlier evaluation.

\begin{table*}[!tbp]
\centering
\caption{Controlled exterior-loss diagnosis with the fixed F-Cooper detector. All counts use score $>0.27$ and BEV overlap $\geq0.7$. In panel (a), Lost and Gained compare exterior removal with Original. Object and Exterior count recoveries among Lost after restoring object-region features or the three rotated exterior regions. Exterior lists $90^\circ/180^\circ/270^\circ$. Stable loss counts new misses among originally correct observations that remain correct after removal. Panel (b) reports all-target counts and global AP70 multiplied by 100 over all 375 inputs.}
\label{tab:supp_fcooper_mechanism}
\small
\setlength{\tabcolsep}{3pt}
\begin{minipage}[t]{.62\textwidth}\centering
(a) Target observations and restoration\par\smallskip
\begin{tabular}{@{}lrrrrrr@{}}
\toprule
Recording & \shortstack{Original\\correct} & Lost & Gained & Object & Exterior & \shortstack{Stable\\loss} \\
\midrule
Dry & 1,166 & 30 & 54 & 30 & 0/0/0 & 0 \\
Rain & 572 & 13 & 16 & 13 & 0/0/0 & 0 \\
All & 1,738 & 43 & 70 & 43 & 0/0/0 & 0 \\
\bottomrule
\end{tabular}
\end{minipage}\hfill
\begin{minipage}[t]{.35\textwidth}\centering
(b) Complete-input performance\par\smallskip
\begin{tabular}{@{}lrrr@{}}
\toprule
Configuration & Correct & False & AP70 \\
\midrule
Original & 1,738 & 2,950 & 25.67 \\
Exterior loss & 1,765 & 2,787 & 28.25 \\
Object restoration & 1,738 & 2,793 & 27.52 \\
\bottomrule
\end{tabular}
\end{minipage}
\end{table*}

Table~\ref{tab:supp_fcooper_mechanism}(a) shows that object-region restoration recovers every induced miss without losing any of the 1,695 stable observations. These misses span 39 physical times and 22 annotated tracks, so the counts include repeated target observations.
To identify detections that depended on cooperation, we examine receiver-only predictions for the same targets. For 20 of the 43 misses, the receiver misses the target before and after exterior removal, with 17 in dry inputs and 3 in rain. The other receiver-only outcomes are 18 correct before and after removal, 3 changing from correct to missed, and 2 changing from missed to correct. Across all inputs, the receiver-only Original baseline has pooled AP70 34.83, above the cooperative Original baseline.

The complete-input results in panel (b) include targets helped by the exterior change. Removal gains 70 correct observations, all lost again after object-region restoration, so restoration lowers AP relative to removal while remaining above Original.
Equal-count random deletion loses 38, 31, and 32 originally correct observations overall. Its rain-recording losses are 18, 10, and 13, compared with 13 under directional removal, showing that the directional pattern is not consistently more harmful there.
Together, these results trace the affected observations through local representations while showing that exterior changes can help other targets. Since both weather groups undergo the same controlled removal, their comparison does not isolate a natural-weather causal effect.

\subsection{Candidate Repair on Original Inputs}

We train on dry inputs and variants with object interiors preserved and exterior points removed from the receiver, peer, or both (six crops per time and condition, 768 patches). The three fits (seeds 20261019--20261021) share inputs and masks but vary initialization, crop sampling, training order and regressor state (Sections~\ref{supp:local_training} and~\ref{supp:quality_training}). We evaluate the final models on unmodified recordings.

\begin{table*}[!tbp]
\centering
\caption{F-Cooper on the same rain and dry inputs with receiver-only detection, cooperative detection (Original), or repair. AWCR uses CoInfra's $\lambda=0.7$. Panel (a) reports Vehicle BEV AP ($\times100$). Panel (b) counts detections at score $>0.27$ and overlap $\geq0.7$. Retained counts target observations detected correctly by both the row configuration and Original. For repair rows, AP is mean$\pm$sample SD and counts are means across three fits. Bold marks the highest AP per recording.}
\label{tab:supp_natural_rain}
\small
\setlength{\tabcolsep}{4pt}
\begin{tabular}{@{}llrrcrrr@{}}
\toprule
& & \multicolumn{2}{c}{(a) Vehicle BEV AP} & & \multicolumn{3}{c}{(b) Fixed-threshold counts} \\
\cmidrule(lr){3-4}\cmidrule(l){6-8}
Recording & Configuration & AP50 & AP70 & & Correct & False & Retained \\
\midrule
Rain (168 times) & Receiver-only & \textbf{59.81} & \textbf{45.04} & & 607 & 638 & 546 \\
& Original & 46.37 & 32.86 & & 572 & 922 & 572 \\
& $\lambda=0$ & 52.78$\pm$0.29 & 37.84$\pm$0.24 & & 577.67 & 772.00 & 556.67 \\
& AWCR & 52.86$\pm$0.69 & 37.86$\pm$0.97 & & 574.67 & 975.00 & 545.00 \\
\midrule
Dry (207 times) & Receiver-only & \textbf{47.17} & \textbf{31.45} & & 1,274 & 1,374 & 1,065 \\
& Original & 38.06 & 23.45 & & 1,166 & 2,028 & 1,166 \\
& $\lambda=0$ & 46.39$\pm$0.19 & 29.57$\pm$0.18 & & 1,242.00 & 1,674.33 & 1,139.33 \\
& AWCR & 45.22$\pm$0.94 & 28.40$\pm$0.65 & & 1,204.67 & 2,064.00 & 1,102.00 \\
\bottomrule
\end{tabular}
\end{table*}

AWCR improves cooperative AP70 on both recordings but falls short of receiver-only detection (Table~\ref{tab:supp_natural_rain}). It adds 53 false detections in rain and 36 in dry conditions. Selection alone ($\lambda=0$) nearly matches AWCR in AP70 in rain and scores higher in dry conditions. It also retains more correct detections and produces fewer false detections on both recordings.

\section{Simulated Weather Input Filtering}
\label{supp:simulation}\label{supp:classifier}\label{supp:pretrained_weather}

In a separate V2X-R experiment, we filter peer returns before detection using Point Classifier (Figure~\ref{fig:framework}(c)). It learns to distinguish original LiDAR returns from weather-added returns using the point labels distributed with V2X-R~\cite{huang2025v2xr,v2xr_model_release}.
We compare the same three detector checkpoints with and without peer filtering on the full test split.

\begin{table*}[!tbp]
\centering
\caption{Point Classifier on the full official V2X-R TEST split (vehicle BEV AP $\times100$). Each weather covers 3,166 times from 16 scenes. Arrows compare the same released checkpoint before and after peer filtering. Weather-specific filters use only the official TRAIN split, and the same three fits are reused across checkpoints. Filtered results are mean$\pm$sample SD. Native AP follows the released frame-order evaluator. Gains are computed before rounding, and bold marks the better result within each pair.}
\label{tab:supp_full_filtering}
\small
\setlength{\tabcolsep}{5pt}
\begin{tabular}{@{}llrrr@{}}
\toprule
Checkpoint & Weather & AP50, Original $\rightarrow$ Filtered & AP70, Original $\rightarrow$ Filtered & $\Delta$AP70 \\
\midrule
CoAlign & Fog & 48.26 $\rightarrow$ \textbf{55.45}$\pm$0.05 & 44.36 $\rightarrow$ \textbf{50.64}$\pm$0.07 & $+6.27$ \\
& Snow & 48.89 $\rightarrow$ \textbf{65.93}$\pm$0.08 & 38.93 $\rightarrow$ \textbf{55.49}$\pm$0.05 & $+16.55$ \\
\shortstack[l]{Clean-trained\\AttFuse} & Fog & 52.56 $\rightarrow$ \textbf{69.31}$\pm$0.09 & 38.34 $\rightarrow$ \textbf{51.52}$\pm$0.17 & $+13.18$ \\
& Snow & 45.88 $\rightarrow$ \textbf{68.05}$\pm$0.08 & 30.51 $\rightarrow$ \textbf{50.70}$\pm$0.04 & $+20.18$ \\
\shortstack[l]{Fog-trained\\AttFuse} & Fog & 78.41 $\rightarrow$ \textbf{83.82}$\pm$0.01 & 61.49 $\rightarrow$ \textbf{68.34}$\pm$0.04 & $+6.86$ \\
& Snow & 65.44 $\rightarrow$ \textbf{74.18}$\pm$0.08 & 45.74 $\rightarrow$ \textbf{54.17}$\pm$0.06 & $+8.43$ \\
\bottomrule
\end{tabular}
\end{table*}

\begin{table}[!tbp]
\centering
\caption{Public LiDAR-only V2X-R weather results and detectors with Point Classifier (AP70 $\times100$). Published rows quote the official V2X-R model release and the LiDAR-only ablation in HRCP Table~4(b). We train Point Classifier on weather-specific TRAIN point labels and select the filter using the known test weather. Filtered results report the mean$\pm$sample SD over three fits per weather, shared across checkpoints. Bold marks the highest AP70 in each weather among the filtered checkpoints and the listed public LiDAR-only results.}
\label{tab:supp_filtering_public}
\small
\setlength{\tabcolsep}{4pt}
\begin{tabular}{@{}lrr@{}}
\toprule
Method / source & Fog & Snow \\
\midrule
\shortstack[l]{Fog-trained AttFuse\\official release~\cite{v2xr_model_release}} & 61.52 & 45.54 \\
HRCP, LiDAR-only ablation~\cite{yang2026hrcp} & 50.28 & 51.75 \\
CoAlign + Point Classifier & 50.64$\pm$0.07 & \textbf{55.49}$\pm$0.05 \\
\shortstack[l]{Fog-trained AttFuse\\+ Point Classifier} & \textbf{68.34}$\pm$0.04 & 54.17$\pm$0.06 \\
\bottomrule
\end{tabular}
\end{table}

\subsection{Filter Implementation and Evaluation}
\label{supp:point_filtering}
\label{sec:point_classifier}
To decide which peer returns to keep, Point Classifier uses six histogram gradient-boosting classifiers $C_k$ to score the descriptor $\phi(x;X_p)$ of each point $x$ in the peer cloud $X_p$.
A classifier votes for weather when its score meets or exceeds $\tau_k$, the 99.5th percentile of its scores on original returns in the calibration scene.
We remove a point when at least four of the six classifiers vote for weather. With $\mathbf{1}[\cdot]$ denoting an indicator, the retention mask $m$ is
\begin{equation}
  \begin{aligned}
    m(x)&=\mathbf{1}\!\left[\sum_{k=1}^{6}
      \mathbf{1}[C_k(\phi(x;X_p))\geq\tau_k]<4\right],\\
    X_p^f&=X_p[m].
  \end{aligned}
  \label{eq:point_mask}
\end{equation}
The mask preserves the values and order of retained peer returns.
The detector receives each filtered peer cloud alongside the unfiltered current receiver cloud.

The first eight descriptor entries are horizontal range, height, intensity, four deviations from the within-cell XYZI medians, and the cell's point count.
Two further entries give the fourth nonzero neighbor distance in XYZI space and the number of neighbors at nonzero distance at most one. If fewer than four neighbors have nonzero distance, we use the farthest of them, or zero if none exists.
The eleventh entry counts matches under a fixed index permutation. For each peer, the occupied-cell keys are stored in $(z,y,x)$ order. For a point in cell $(z,y,x)$, the implementation counts how many keys $(x,z+u,y+v)$ occur in that set over $u,v\in\{-1,0,1\}$ with $(u,v)\ne(0,0)$. Fitting and inference use this same lookup.

For training, simulator labels assign zero to original points and one to added returns.
In the full-test study, we train separate fog and snow ensembles using eight fixed times from each of six official TRAIN scenes, or 48 times per weather.
The snow ensemble therefore uses TRAIN snow labels.
In fold $j$, scene $j$ is held out, scene $(j+1)\bmod6$ provides calibration points, and the remaining four scenes provide fitting points.
Each classifier therefore uses 32 fitting times and eight calibration times, with up to 100 iterations, 15 leaves, a learning rate of 0.05, and deterministic caps of 200,000 points per class.
We repeat fitting with seeds 177, 178, and 179 while keeping the selected times, folds, calibration rule, and voting rule fixed.
For each weather and seed, all detection checkpoints receive the same trained filter, including the fog-trained AttFuse checkpoint evaluated below.
Its calibration thresholds and voting rule stay unchanged across checkpoints. At test time, the known simulated weather selects the corresponding ensemble.

\begin{table*}[!tbp]
\centering
\caption{Detection retention and all-annotation AP on the full V2X-R TEST split. (a) Counts use native predictions above score 0.20 and BEV overlap 0.7. Original correct and Original false are single fixed-detector counts, while filtered counts are three-fit means. Retained divides the mean retained count by Original correct. (b) Both runs in each row use all eligible annotations as their common recall denominator, including those in frames with \texttt{None} output. Predictions and ordering match the native AP evaluation in Table~\ref{tab:supp_full_filtering}.}
\label{tab:supp_filtering_retention}
\small
\setlength{\tabcolsep}{5pt}
(a) Changes in target detections\par\smallskip
\begin{tabular}{@{}llrrrrr@{}}
\toprule
Checkpoint & Weather & Original correct & Retained (\%) & Gained & Lost & False, Original $\rightarrow$ Filtered \\
\midrule
CoAlign & Fog & 12,321 & 99.63 & 2,312.33 & 45.33 & 1,698 $\rightarrow$ 2,343.00 \\
& Snow & 12,119 & 97.20 & 5,446.33 & 339.67 & 3,840 $\rightarrow$ 4,265.33 \\
\shortstack[l]{Clean-trained\\AttFuse} & Fog & 7,919 & 98.29 & 3,710.67 & 135.67 & 1,991 $\rightarrow$ 3,074.00 \\
& Snow & 10,513 & 96.03 & 6,246.33 & 417.00 & 5,023 $\rightarrow$ 5,484.00 \\
\shortstack[l]{Fog-trained\\AttFuse} & Fog & 15,393 & 99.01 & 494.00 & 153.00 & 5,531 $\rightarrow$ 4,477.00 \\
& Snow & 15,674 & 96.33 & 2,722.33 & 575.33 & 6,581 $\rightarrow$ 6,585.67 \\
\bottomrule
\end{tabular}
\par\smallskip
(b) AP70 with all eligible annotations in the denominator\par\smallskip
\begin{tabular}{@{}llrrrr@{}}
\toprule
Checkpoint & Weather & Eligible annotations & Original & Filtered & $\Delta$AP70 \\
\midrule
CoAlign & Fog & 26,153 & 43.71 & 50.01$\pm$0.06 & $+6.31$ \\
& Snow & 26,673 & 37.12 & 55.28$\pm$0.05 & $+18.16$ \\
\shortstack[l]{Clean-trained\\AttFuse} & Fog & 18,518 & 35.81 & 50.12$\pm$0.16 & $+14.31$ \\
& Snow & 25,956 & 29.08 & 50.36$\pm$0.05 & $+21.28$ \\
\shortstack[l]{Fog-trained\\AttFuse} & Fog & 18,518 & 61.45 & 68.30$\pm$0.04 & $+6.85$ \\
& Snow & 25,956 & 45.68 & 54.10$\pm$0.06 & $+8.42$ \\
\bottomrule
\end{tabular}
\end{table*}

\subsection{Full-Test Results and Public References}
\label{supp:full_filtering}
We evaluate all 3,166 official TEST times from 16 scenes for each weather and detector.
CoAlign and clean-trained AttFuse use their released LiDAR checkpoints.
To test whether filtering remains useful after weather adaptation, we also evaluate the official fog-trained AttFuse checkpoint, which was trained on clear and fog inputs~\cite{v2xr_model_release}.
All checkpoints retain their native preprocessing, field-of-view limits, score threshold of 0.20, and suppression rules.
The filter acts on peer points retained by native preprocessing before feature encoding, keeping receiver inputs unchanged and preserving retained peer values and order.
The released CoAlign configuration accepts annotated targets visible to LiDAR or radar, whereas both AttFuse checkpoints require LiDAR visibility.
For each checkpoint, the original and filtered runs use the same test frames and annotation-eligibility rules. The detectors and Point Classifier use only LiDAR inputs.

Table~\ref{tab:supp_full_filtering} reports native AP. The released evaluation code sorts predictions within each frame and accumulates them in frame order~\cite{v2xr_model_release}. It skips a frame and its annotations when the detector returns \texttt{None}, but counts the annotations for an empty prediction array. Filtering can therefore change the native recall denominator.
Table~\ref{tab:supp_filtering_retention}(b) fixes the recall denominator within each checkpoint--weather pair to all eligible annotations across the test frames. Predictions, ordering, and precision are identical to those used for native AP. For CoAlign in snow, the AP70 gain is 16.55 points under native evaluation and 18.16 points with all 26,673 eligible annotations counted.

Input filtering improves AP70 for every checkpoint in both weather conditions on the full test split (Table~\ref{tab:supp_full_filtering}).
With fog-trained AttFuse, the same filters raise fog AP70 from 61.49 to $68.34\pm0.04$, showing additional benefit after weather adaptation.
For CoAlign in snow, AP70 rises from 38.93 to $55.49\pm0.05$.

Table~\ref{tab:supp_filtering_public} compares the filtered detectors with the official fog-trained AttFuse release and the LiDAR-only ablation reported in HRCP.
The snow filter uses point labels from TRAIN snow inputs, while the released fog-trained detector was trained on clear and fog inputs.
The paired gains in Table~\ref{tab:supp_full_filtering} measure the additional benefit of filtering with each detector held fixed.

To examine what happens to individual detections, we track each annotated target at each test time.
Table~\ref{tab:supp_filtering_retention}(a) uses native predictions after the released 0.20 score threshold and suppression, with matches at BEV overlap 0.7.
Retained gives the fraction of originally correct observations still detected after filtering. Gained and Lost count newly correct observations and originally correct observations that become misses, while False counts unmatched predictions, including duplicates and inaccurate localization.
With these definitions, the filter retains between 96.03\% and 99.63\% of originally correct observations while recovering additional targets in every setting.
On fog-trained AttFuse, filtering also reduces fog false detections from 5,531 to 4,477, with 99.01\% retention of originally correct observations, and leaves the snow false-detection count nearly unchanged.
For the clean-trained checkpoints, recovery instead accompanies higher false-detection counts.
Despite these different count tradeoffs, AP70 gains remain positive in all six settings when every eligible annotation contributes to recall (Table~\ref{tab:supp_filtering_retention}(b)).

We also check preservation on the full clean TEST split by applying both weather filters, without selecting a filter from these test results.
The fog filter leaves clean AP70 and aggregate detection counts unchanged across all checkpoints and fits.
For fog-trained AttFuse, clean AP70 remains at 76.09.
The snow filter changes CoAlign clean AP70 from 80.04 to $79.55\pm0.05$, clean-trained AttFuse from 80.05 to $79.81\pm0.05$, and fog-trained AttFuse from 76.09 to $75.80\pm0.09$.
These small declines accompany retention of 99.62\%, 99.54\%, and 99.50\% of originally correct clean observations, respectively.

On six fog inputs, filtering raised median FP32 processing time from 317 to 1,454\,ms under the timing protocol in Section~\ref{supp:cost}, used 1,384\,MiB of peak allocated GPU memory, and reduced median peer FP32 XYZI payload from 0.42 to 0.39\,MiB before transport overhead.

\clearpage
\newcommand{\ControlledAccurateRows}{%
1739396553.2 & 7 & 740 & 0.7911 & \texttt{OOO} & \texttt{OOO} & \texttt{OOO} \\
1739396553.2 & 8 & 739 & 0.8300 & \texttt{OOO} & \texttt{OBB} & \texttt{OBB} \\
1739396559.0 & 7 & 733 & 0.7470 & \texttt{OOO} & \texttt{OOO} & \texttt{OOO} \\
1739396564.8 & 7 & 738 & 0.7087 & \texttt{OOO} & \texttt{OOO} & \texttt{OOO} \\
1739396632.0 & 8 & 747 & 0.8827 & \texttt{OOO} & \texttt{OOO} & \texttt{OOO} \\
1739396634.3 & 6 & 753 & 0.7652 & \texttt{OOO} & \texttt{OOO} & \texttt{OOO} \\
1739396634.3 & 8 & 759 & 0.7051 & \texttt{OOO} & \texttt{OOO} & \texttt{OOO} \\
1739396652.8 & 7 & 754 & 0.8313 & \texttt{OBB} & \texttt{OBB} & \texttt{OBB} \\
1742850631.1 & 8 & 45 & 0.7044 & \texttt{OOO} & \texttt{OOO} & \texttt{OOO} \\
1742850631.1 & 8 & 47 & 0.9447 & \texttt{BBB} & \texttt{BBB} & \texttt{BBB} \\
1742850639.3 & 7 & 47 & 0.7468 & \texttt{OOO} & \texttt{OOO} & \texttt{OOO} \\
1742850639.3 & 8 & 47 & 0.7935 & \texttt{OOO} & \texttt{OOO} & \texttt{OOO} \\
1742850765.2 & 8 & 73 & 0.7082 & \texttt{OOO} & \texttt{OOO} & \texttt{OOO} \\
1742850773.8 & 7 & 76 & 0.8553 & \texttt{OOO} & \texttt{OOO} & \texttt{OOO} \\
1742850782.5 & 8 & 77 & 0.7107 & \texttt{OOO} & \texttt{OOO} & \texttt{OOO} \\
1743109106.5 & 7 & 137 & 0.8377 & \texttt{OOO} & \texttt{-OO} & \texttt{OBB} \\
1743109106.5 & 7 & 141 & 0.8373 & \texttt{OBO} & \texttt{OBO} & \texttt{OBO} \\
1743109109.3 & 7 & 145 & 0.7287 & \texttt{OOO} & \texttt{OOO} & \texttt{OOO} \\
1743109109.3 & 9 & 136 & 0.7540 & \texttt{OBB} & \texttt{BBB} & \texttt{OBB} \\
1743109117.9 & 7 & 137 & 0.7574 & \texttt{OOO} & \texttt{OOO} & \texttt{OBB} \\
1743109117.9 & 8 & 137 & 0.7319 & \texttt{OOO} & \texttt{OOO} & \texttt{OOO} \\
1743627904.4 & 8 & 515 & 0.7018 & \texttt{OOO} & \texttt{OOO} & \texttt{OOO} \\
1743627940.4 & 7 & 521 & 0.7619 & \texttt{OBB} & \texttt{OBB} & \texttt{OBB} \\
1743627940.4 & 8 & 521 & 0.7325 & \texttt{O--} & \texttt{O--} & \texttt{O--} \\
1744060679.8 & 8 & 629 & 0.7369 & \texttt{BOO} & \texttt{OOO} & \texttt{OOO} \\
1744060682.0 & 7 & 632 & 0.7017 & \texttt{OOO} & \texttt{OOO} & \texttt{OOO} \\
1763674905.0 & 8 & 1406 & 0.7293 & \texttt{BOO} & \texttt{BOO} & \texttt{BOO} \\
1763674922.1 & 8 & 1409 & 0.7595 & \texttt{OOO} & \texttt{OOO} & \texttt{O--} \\
1763674927.9 & 7 & 1422 & 0.7159 & \texttt{OOO} & \texttt{OOO} & \texttt{OOO} \\
1763674939.3 & 9 & 1415 & 0.7814 & \texttt{OOO} & \texttt{OOO} & \texttt{OOO} \\
1763674945.0 & 8 & 1421 & 0.8286 & \texttt{BBB} & \texttt{BOO} & \texttt{BOO} \\%
}

\newcommand{\ControlledNoAccurateRows}{%
1739396564.8 & 8 & 738 & 0.6877 & \texttt{OOO} & \texttt{OOO} & \texttt{OOO} \\
1739396616.8 & 6 & 743 & 0.6767 & \texttt{OOO} & \texttt{OOO} & \texttt{OOO} \\
1739396626.9 & 8 & 742 & 0.6977 & \texttt{OOO} & \texttt{OOO} & \texttt{OOO} \\
1739396634.3 & 7 & 759 & 0.5851 & \texttt{OOO} & \texttt{OOO} & \texttt{OOO} \\
1739396652.8 & 7 & 761 & 0.6558 & \texttt{OOO} & \texttt{OOO} & \texttt{OOO} \\
1742850631.1 & 7 & 45 & 0.6621 & \texttt{OOO} & \texttt{OOO} & \texttt{OOO} \\
1742850631.1 & 7 & 46 & 0.6432 & \texttt{OOO} & \texttt{OOO} & \texttt{OOO} \\
1742850631.1 & 8 & 46 & 0.6128 & \texttt{OOO} & \texttt{OOO} & \texttt{OOO} \\
1742850698.7 & 7 & 55 & 0.6072 & \texttt{OOO} & \texttt{OOO} & \texttt{OOO} \\
1742850698.7 & 8 & 55 & 0.6666 & \texttt{OOO} & \texttt{OOO} & \texttt{OOO} \\
1742850698.7 & 8 & 56 & 0.6086 & \texttt{OOO} & \texttt{OOO} & \texttt{OOO} \\
1742850756.5 & 8 & 71 & 0.6129 & \texttt{OOO} & \texttt{OOO} & \texttt{OOO} \\
1742850765.2 & 7 & 73 & 0.6282 & \texttt{OOO} & \texttt{OOO} & \texttt{OOO} \\
1742850782.5 & 7 & 77 & 0.6765 & \texttt{OOO} & \texttt{OOO} & \texttt{OOO} \\
1743109100.8 & 7 & 134 & 0.6554 & \texttt{OOO} & \texttt{OOO} & \texttt{OOO} \\
1743109100.8 & 8 & 134 & 0.6689 & \texttt{OOO} & \texttt{OOO} & \texttt{OOO} \\
1743109103.6 & 8 & 141 & 0.6526 & \texttt{OOO} & \texttt{OOO} & \texttt{OOO} \\
1743109103.6 & 9 & 134 & 0.6929 & \texttt{O--} & \texttt{O--} & \texttt{OD-} \\
1743109106.5 & 7 & 145 & 0.6791 & \texttt{OOO} & \texttt{OOO} & \texttt{OOO} \\
1743109106.5 & 8 & 137 & 0.6206 & \texttt{OOO} & \texttt{OOO} & \texttt{OOO} \\
1743109115.0 & 9 & 145 & 0.6473 & \texttt{OOO} & \texttt{OOO} & \texttt{OOO} \\
1743627760.0 & 8 & 465 & 0.6826 & \texttt{OOO} & \texttt{OOO} & \texttt{OOO} \\
1743627928.4 & 9 & 522 & 0.6637 & \texttt{OOO} & \texttt{OOO} & \texttt{OOO} \\
1743628133.4 & 8 & 553 & 0.6915 & \texttt{OOO} & \texttt{OOO} & \texttt{OOO} \\
1744060673.3 & 7 & 628 & 0.6937 & \texttt{OOO} & \texttt{OOO} & \texttt{OOO} \\
1744060673.3 & 8 & 628 & 0.6409 & \texttt{OOO} & \texttt{OOO} & \texttt{OOO} \\
1744060675.5 & 7 & 626 & 0.6973 & \texttt{OOO} & \texttt{OOO} & \texttt{OOO} \\
1744060675.5 & 8 & 625 & 0.6936 & \texttt{OOO} & \texttt{OOO} & \texttt{OOO} \\
1744060677.7 & 7 & 625 & 0.6287 & \texttt{OOO} & \texttt{OOO} & \texttt{OOO} \\
1744060677.7 & 7 & 628 & 0.6676 & \texttt{OOO} & \texttt{OOO} & \texttt{OOO} \\
1744060677.7 & 8 & 628 & 0.5998 & \texttt{---} & \texttt{---} & \texttt{---} \\
1744060679.8 & 7 & 629 & 0.6956 & \texttt{OOO} & \texttt{OOO} & \texttt{OOO} \\
1744060679.8 & 8 & 631 & 0.6289 & \texttt{OOO} & \texttt{OOO} & \texttt{OOO} \\
1744060682.0 & 7 & 631 & 0.6510 & \texttt{OOO} & \texttt{OOO} & \texttt{OOO} \\
1744060682.0 & 8 & 631 & 0.5707 & \texttt{OOO} & \texttt{OOO} & \texttt{OOO} \\
1744060682.0 & 8 & 632 & 0.6870 & \texttt{OOO} & \texttt{OOO} & \texttt{OOO} \\
1744060684.2 & 7 & 633 & 0.6607 & \texttt{OOO} & \texttt{OOO} & \texttt{OOO} \\
1744060684.2 & 8 & 633 & 0.6605 & \texttt{OOO} & \texttt{OOO} & \texttt{OOO} \\
1763674910.7 & 7 & 1405 & 0.4053 & \texttt{OOO} & \texttt{OOO} & \texttt{OOO} \\
1763674916.4 & 8 & 1429 & 0.5963 & \texttt{OOO} & \texttt{OOO} & \texttt{OOO} \\
1763674922.1 & 8 & 1429 & 0.5668 & \texttt{OOO} & \texttt{OOO} & \texttt{OOO} \\%
}

\begin{table*}[p]
\section{Target-Level Outcomes of Controlled Candidate Repair}
\label{supp:controlled_target_rows}
\centering
\caption{Target-by-target outcomes for the 72 CoInfra misses induced by exterior removal (Section~\ref{supp:controlled_candidates}). Each observation is identified by the dataset timestamp, receiving node (Rx), and vehicle annotation ID. IoU is the best overlap of any valid original candidate after deletion. For each fit, the three symbols give the outcomes of original-candidate reranking, full AWCR, and fixed-score coordinate rollback, in that order. \texttt{B} means detected on both inputs, \texttt{O} on the original input only, \texttt{D} on the exterior-deleted input only, and \texttt{-} on neither. Fits 1--3 follow Table~\ref{tab:supp_controlled_bridge}.}
\label{tab:supp_controlled_target_rows}
\small
\setlength{\tabcolsep}{2.5pt}
\begin{minipage}[t]{.485\textwidth}
\centering
Accurate original candidate available (31)\par\smallskip
\begin{tabular}{@{}rrrrccc@{}}
\toprule
Timestamp & Rx & ID & IoU & Fit 1 & Fit 2 & Fit 3 \\
\midrule
\ControlledAccurateRows
\bottomrule
\end{tabular}
\end{minipage}\hfill
\begin{minipage}[t]{.485\textwidth}
\centering
No accurate original candidate (41)\par\smallskip
\begin{tabular}{@{}rrrrccc@{}}
\toprule
Timestamp & Rx & ID & IoU & Fit 1 & Fit 2 & Fit 3 \\
\midrule
\ControlledNoAccurateRows
\bottomrule
\end{tabular}
\end{minipage}
\end{table*}

\end{document}